\documentclass{article} 
\usepackage{iclr2027_conference,times}

\usepackage{amsmath,amsfonts,bm}

\def\eqref#1{equation~\ref{#1}}

\def\1{\bm{1}}

\DeclareMathAlphabet{\mathsfit}{\encodingdefault}{\sfdefault}{m}{sl}
\SetMathAlphabet{\mathsfit}{bold}{\encodingdefault}{\sfdefault}{bx}{n}

\usepackage{hyperref}
\usepackage{url}
\usepackage{amsmath}
\usepackage{subcaption}
\usepackage{float}
\usepackage{graphicx}
\usepackage{verbatim}
\usepackage[utf8]{inputenc} 
\usepackage[T1]{fontenc}    
\usepackage{textcomp} 
\usepackage{makecell} 
\usepackage{threeparttable}
\usepackage{multirow}
\usepackage{cleveref}
\usepackage{booktabs}       
\usepackage{amsfonts}       
\usepackage{nicefrac}       
\usepackage{microtype}      
\usepackage{xcolor}         
\usepackage{enumitem}

\newcommand{\ppm}[1]{\,\scriptsize{$\pm$#1}}
\newcommand{\mypar}[1]{\vspace{1pt}\noindent\textbf{#1}~}

\title{Latent Goal Prediction from Language for Model-Based Planning}

\author{%
  \begin{minipage}{\textwidth}
  \centering
  \textbf{Samuel Barbeau}$^{1,2,\dagger}$\thanks{%
    $^{1}$École de Technologie Supérieure, Montréal \quad
    $^{2}$ILLS, CNRS, Montreal, QC H2S 3H1 \quad
    $^{3}$Polytechnique Montréal \quad
    $^{4}$MILA \quad
    $^{5}$Institut Universitaire de France (IUF) \quad
    $^{6}$Sorbonne Université, CNRS, ISIR Paris \\[0.3ex]
    $^{*}$Equal advising \quad
    $^{\dagger}$Corresponding author: \texttt{samuel.barbeau.2@ens.etsmtl.ca}%
  } \quad
  \textbf{Simon Roy}$^{3,4}$ \quad
  \textbf{Giovanni Beltrame}$^{3,4}$ \\[0.5ex]
  \textbf{Christian Desrosiers}$^{1,2,*}$ \quad
  \textbf{Nicolas Thome}$^{2,5,6,*}$
  \end{minipage}
}

\iclrfinalcopy 
\begin{document}
\maketitle
\lhead{Preprint.}

\begin{abstract}
Joint-Embedding Predictive Architectures (JEPAs) enable agents to plan in latent space by imagining the outcomes of candidate actions, yet task specification remains a bottleneck. Visual targets provide precise
local gradients but poor distant guidance, while language is flexible yet limited by noisy cross-modal alignment or dependence on distinct large generative models. We introduce LAGO (Latent Goal Prediction from Language), a hierarchical world model in which a single predictor both forecasts action-conditioned dynamics and grounds language instructions as sequences of intermediate latent subgoals, training both modes with a single regression objective over a shared latent space. At each planning step, LAGO predicts a sequence of latent subgoals from a language instruction and optimizes an action sequence using a soft-minimum alignment cost that rewards subgoal proximity without enforcing a rigid path. Subgoals are repredicted as the agent's state evolves, turning a single long-horizon instruction into a sequence of locally tractable objectives. Across environments spanning navigation and manipulation, LAGO outperforms flat and hierarchical image-goal planners, with the largest gains on long-horizon navigation with curved, hazard-constrained paths, where it more than doubles the success rate of a goal-conditioned behavior cloning policy trained on the same demonstrations. Grounding language in the world model's own latent space also outperforms vision-language reward models, including one that plans with ground-truth dynamics.
\end{abstract}

\begin{figure}[hb!]
    \centering
    \includegraphics[width=0.92\textwidth]{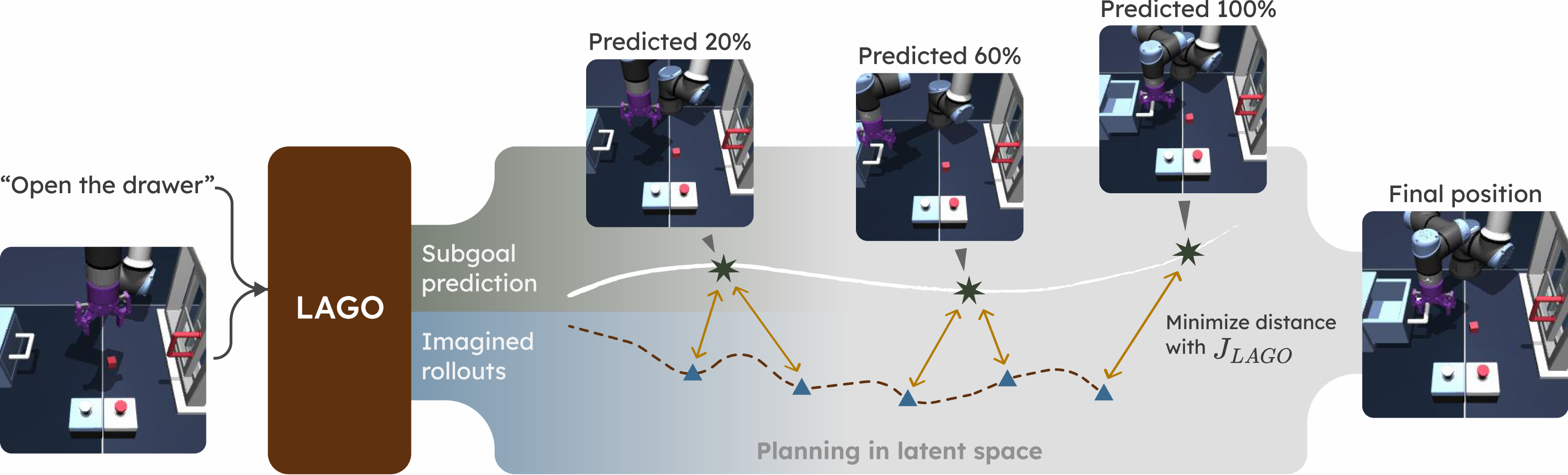} 
    \caption{\textbf{Latent Goal Prediction from Language for Model-Based Planning.} LAGO trains a single latent predictor to forecast both language-conditioned subgoals (green stars) and action-conditioned state rollouts (blue triangles), unifying task specification and dynamics in a shared latent space. Given a language instruction, CEM optimizes action sequences to reach the predicted subgoals using $J_{\text{LAGO}}$. Re-predicting subgoals from the agent's current state at each planning step enables adaptive, long-horizon planning toward the desired final state.}
    \label{fig:lago_high_level}
    \vspace{-10pt}
\end{figure}

\section{Introduction}
\label{introduction}
\vspace*{-5pt}

Achieving complex tasks from high-dimensional sensory observations remains a fundamental challenge in embodied intelligence. An agent must learn how its actions affect the world while reasoning over long horizons to reach a desired outcome. World models \citep{hou2026worldmodelrobotlearning} offer a promising framework for this problem by separating these two challenges: a model can learn the dynamics of the environment and then be used to plan action sequences through imagined interactions, rather than learning a direct mapping from observations to actions.
Two broad approaches have emerged for learning such models. One line of work jointly learns the environment dynamics with task-specific rewards or value functions, allowing planning to optimize a learned notion of task progress \citep{sutton2018reinforcement,hafner2020dream,hansen2022tdmpc}. While such methods have demonstrated strong data efficiency and multitask capabilities \citep{hafner2025mastering}, they require additional reward supervision and are therefore tied to the tasks for which such supervision is available. A second line of work instead learns the dynamics of the environment without specifying a task at training time. In particular, Joint-Embedding Predictive Architectures (JEPAs) \citep{lecun2022path} learn to predict future representations from observations and actions through a self-supervised latent-regression objective. V-JEPA 2 \citep{assran2025vjepa2selfsupervisedvideo} demonstrated the potential of this paradigm for learning visual dynamics at scale, while subsequent work has leveraged such representations for trajectory optimization in robotic environments \citep{zhou2025dinowmworldmodelspretrained,daniel2026latentparticleworldmodels,zhang2026hierarchicalplanninglatentworld,maes2026leworldmodelstableendtoendjointembedding,masip2026ffjepalonghorizonplanningworld}.

Existing JEPA-based planners formulate the desired outcome as a goal image, optimizing an action sequence such that its predicted latent state approaches the representation of the target~\citep{zhou2025dinowmworldmodelspretrained}. Such a goal image provides precise supervision and a well-defined objective, but is impractical to obtain and becomes limiting for long-horizon tasks when the target lies beyond the model's reliable prediction horizon. Recent work mitigates this either by encouraging straighter, more predictable latent trajectories~\citep{wang2026temporalstraighteninglatentplanning,maes2026leworldmodelstableendtoendjointembedding}, or by decomposing the horizon hierarchically, predicting intermediate subgoals or planning across multiple temporal scales~\citep{zhang2026hierarchicalplanninglatentworld}. Despite these advances, all of these methods still require a visual goal to specify the task.

Language offers a flexible alternative for specifying tasks, allowing a user to describe the desired behavior rather than provide a target observation. The challenge is then to ground this abstract instruction in the physical state of the environment. Existing approaches have explored this through language-conditioned rewards, either by training completion classifiers \citep{nair2021learninglanguageconditionedrobotbehavior} or leveraging pretrained vision-language models to score the alignment between language and visual states or trajectories \citep{nair2022r3muniversalvisualrepresentation,sontakke2023roboclipdemonstrationlearnrobot,ma2023livlanguageimagerepresentationsrewards,rocamonde2024visionlanguagemodelszeroshotreward,wang2024rlvlmf}. However, these cross-modal objectives can be noisy and difficult to optimize \citep{fu2024furlvisuallanguagemodelsfuzzy,roy2025revisitinglearningobjectivesvisionlanguage}. Other methods instead predict future visual states from language, using image-editing diffusion models to generate subgoals \citep{black2023zeroshotroboticmanipulationpretrained} or video generation models to predict trajectories and intermediate states \citep{du2023learninguniversalpoliciestextguided,ajay2023compositionalfoundationmodelshierarchical,ko2023learningactactionlessvideos,du2023videolanguageplanning}. These approaches rely on large generative models and expensive pixel-space sampling, while their visual predictions remain separate from the latent space in which the world model learns dynamics, preventing goals and dynamics from being expressed in a common representation.

We introduce \textbf{LAGO} (\textit{Latent Goal prediction from language}), a hierarchical world model that plans directly from natural-language instructions. As presented in Figure~\ref{fig:lago_high_level}, a single predictor operates in two modes: forecasting state dynamics conditioned on action, and grounding a language instruction as latent subgoals in the current scene. These language-conditioned subgoals give planning concrete, reachable targets, turning a long-horizon instruction into a sequence of locally tractable objectives that can be repredicted at every planning step. We summarize our contributions as follows:
\begin{itemize}[itemsep=2pt,topsep=2pt,leftmargin=18pt]
    \item To bridge the gap between image-based and language-based goal specification, we introduce LAGO, a world model trained to map natural language instructions to target states within its own latent space, combining the precision of visual goals with the flexibility of language.
    \item Conditioned on language and an arbitrary completion value, LAGO is capable of predicting many subgoals for one instruction. Coupled with a soft-minimum cost function, this creates a navigable valley guiding the action-conditioned rollouts towards the objective.
    \item We show that planning with subgoals inferred from language approaches the performance of oracle subgoals at short and medium horizons and substantially outperforms flat and hierarchical image-goal planners. Additionally, grounding language in the world model's own latent space outperforms dedicated vision-language reward models, including when the latter plan with ground-truth dynamics.
\end{itemize}

\vspace*{-5pt}
\section{Related Work}
\label{related_work}
\vspace*{-5pt}

\mypar{World Models.} WMs were popularized in the context of model-based reinforcement learning (MBRL) \citep{sutton2018reinforcement, https://doi.org/10.5281/zenodo.1207631}, where an agent learns a model of its environment through which it can plan a policy before executing it. These methods learn latent dynamics from interaction, and  address goal specification by learning task-specific reward and value functions from experience \citep{hafner2019planet,hafner2020dream,hansen2022tdmpc}. The method proposed by \cite{hafner2025mastering} was the first  to collect diamonds in Minecraft without human data or task-specific tuning, and recent work has scaled MBRL across a large and heterogeneous suite of environments \citep{hansen2025learningmassivelymultitaskworld}. Yet, these methods remain bottlenecked not by model capacity but by reward specification. A parallel line of work instead formulates planning as optimizing actions to minimize the distance between predicted rollouts and a goal state in a learned latent space, eliminating the need for task-specific reward and value learning. The representation is either inherited from a pretrained vision encoder~\citep{zhou2025dinowmworldmodelspretrained} or learned from pixels end to end~\citep{assran2025vjepa2selfsupervisedvideo}. 

To improve long-horizon planning, the latent space can be explicitly regularized so that distance in the space approximates planning cost~\citep{wang2026temporalstraighteninglatentplanning}. \cite{maes2026leworldmodelstableendtoendjointembedding} shows that a similar straightening of latent trajectories emerges naturally from end-to-end next-state prediction training, without an explicit temporal regularizer. Long horizons have also been addressed through hierarchical planning, either by predicting intermediate subgoals conditioned on a goal image~\citep{he2026demojepajointembeddingpredictivearchitecture} or by learning world models at different temporal scales, with a high-level model generating intermediate states for a low-level planner~\citep{zhang2026hierarchicalplanninglatentworld}. Common to all these methods is the use of a goal image, which gives precise, well-shaped objectives near the target, but is impractical to obtain for complex tasks and offers little guidance when the goal lies beyond the planning horizon.

LAGO addresses this limitation by automatically generating latent subgoals that decompose long-horizon planning into a sequence of short-horizon prediction problems, substantially reducing the rollout lengths required to obtain meaningful optimization signals. Rather than optimizing directly toward a distant objective, LAGO \emph{dynamically predicts intermediate latent targets} that maintain local tractability while preserving long-horizon semantic consistency.

\mypar{Language as Task Specification.} Vision-language-action (VLA) models map observations and instructions directly to actions by scaling behavior cloning \citep{brohan2023rt2,intelligence2025pi05visionlanguageactionmodelopenworld}, but they must implicitly infer the intermediate behaviors of a long-horizon instruction and scale only with action-labeled data. Reward-based methods instead ground language in a reward signal: \cite{nair2021learninglanguageconditionedrobotbehavior} trains a language-conditioned completion classifier for model-based planning, and later work repurposes pretrained VLMs as rewards via state--language embedding similarity or trajectory preference labels \citep{nair2022r3muniversalvisualrepresentation,sontakke2023roboclipdemonstrationlearnrobot,rocamonde2024visionlanguagemodelszeroshotreward}. This removes hand-designed rewards but yields brittle, noisy objectives \citep{fu2024furlvisuallanguagemodelsfuzzy,roy2025revisitinglearningobjectivesvisionlanguage}. A third line grounds language in future states, using image-editing or video generation models to synthesize subgoals or goal-reaching trajectories, with actions recovered by inverse dynamics \citep{black2023zeroshotroboticmanipulationpretrained,du2023learninguniversalpoliciestextguided,ajay2023compositionalfoundationmodelshierarchical}. Because supervision is a future state, this scales to action-less video datasets orders of magnitude larger than action-labeled ones \citep{grauman2022ego4dworld3000hours,chen2026action100mlargescalevideoaction}, but tying planning to generative pixel decoding is expensive and forfeits scoring trajectories in the space where dynamics are modeled.

LAGO instead maps language instructions directly to target states \emph{within a world model's latent space} and plans by minimizing latent distance to these targets. Because supervision is a future latent state, LAGO avoids the noisy cross-modal objectives of VLM-based rewards and the cost of decoding imagined rollouts into pixels, requiring action labels only to fit a task-agnostic world model and remaining free to scale with action-less demonstration data.

\mypar{Hierarchical Planners.} A growing line of work addresses long-horizon planning by decomposing tasks into intermediate subgoals that are easier to optimize and execute.
An early approach by \cite{nair2019hierarchicalforesightselfsupervisedlearning} trains a generative model to propose candidate subgoal images and optimizes these subgoals to minimize the worst-case planning cost across segments. However, the subgoals are decoded as full images, their number is limited by the cost of optimization, and they are computed once before execution rather than updated as the plan unfolds.
\cite{chen2024simple} generates coarse high-level trajectories subsequently refined into executable plans, while \cite{hao2025chd} introduces coupled diffusion processes to improve temporal consistency. However, these methods plan over low-level environment states, rather than high-dimensional observations, and rely on behavior cloning to execute the resulting plans. \cite{huang2024subgoal} uses a diffusion model to generate coarse-to-fine latent subgoal sequences that are executed with model predictive control (MPC), enabling iterative replanning rather than a fixed hierarchical decomposition. However, it assumes access to privileged geometric states, namely signed distance fields and full 3D object models, which restricts its application in standard visual environments. More recently, \cite{haramati2026hierarchical} explored object-centric subgoal decomposition for compositional planning, but relies on a separately trained value function to filter generated subgoals for reachability at test time. Concurrently, FF-JEPA \citep{masip2026ffjepalonghorizonplanningworld} also keeps subgoal prediction in the world model's latent space, pairing an action-conditioned dynamics model with an action-free planner that proposes the next subgoal for a CEM optimizer. LAGO differs from FF-JEPA in three ways:
\textbf{(i)} it uses a \emph{single predictor} with shared parameters and a unified training objective for both dynamics and subgoal prediction, rather than relying on separate models; \textbf{(ii)} it conditions each subgoal on a language instruction and a continuous completion variable $\rho$ to predict \emph{task-specific} intermediate states; \textbf{(iii)} at test time, it predicts an arbitrary number of subgoals spanning the whole trajectory and scores rollouts with a soft-minimum cost that rewards only their closest approach to each one, so near subgoals act as targets and distant ones as directional guidance. This lets the world model find its own optimal path through them, rather than directly minimizing distance to a single subgoal.

Unlike these approaches, LAGO doesn't require reward supervision, expert action supervision, access to low-level state information and can predict task-specific intermediate states. Additionally, it can predict an arbitrary number of subgoals and continuously update its latent subgoal predictions online as the world state and language instruction evolve. Combined with our soft minimum trajectory cost formulation, which enables smooth progress evaluation without brittle discrete completion criteria, this yields more robust long-horizon planning under model uncertainty.

\vspace*{-5pt}
\section{Methodology}
\label{sec:methodology}
\vspace*{-5pt}

\subsection{LAGO}
\label{training_lago}
\begin{figure}[t!]
    \centering
    \includegraphics[width=\textwidth]{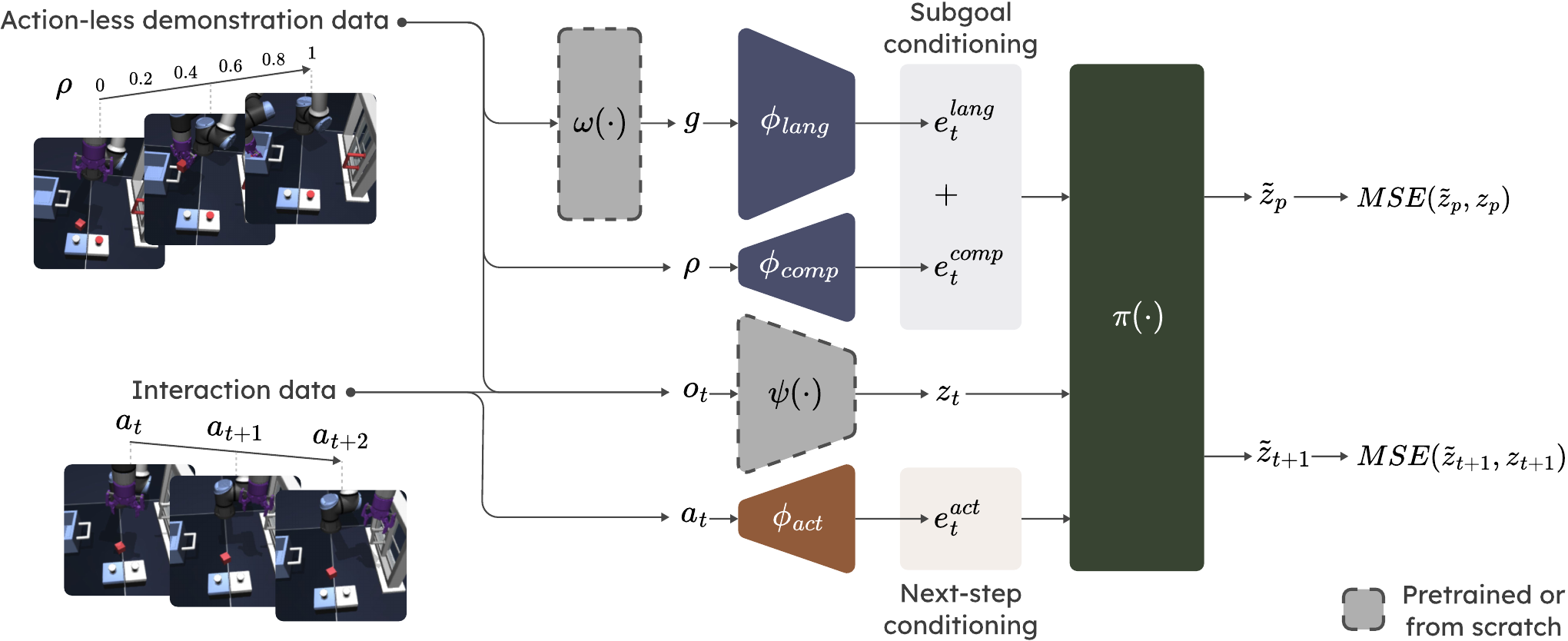} 
    \caption{\textbf{The LAGO architecture.} LAGO is a Joint Embedding Predictive Architecture operating in two modes over a shared latent space. Given a current latent observation $z_t$, a conditioned latent predictor $\pi(\cdot)$ forecasts a target latent state: either the next state $\tilde{z}_{t+1}$ when conditioned on a low-level action, or a subgoal state $\tilde{z}_{\rho}$ when conditioned on a language instruction and a completion scalar $\rho$. Both modes share a single predictor, and the architecture is agnostic to the choice of latent space.}
    \label{fig:architecture}
    \vspace{-10pt}
\end{figure}

LAGO is a Joint Embedding Predictive Architecture \citep{assran2023selfsupervisedlearningimagesjointembedding} supporting two operating modes: \textbf{(1)} one-step transition prediction conditioned on actions, and \textbf{(2)} subgoal prediction conditioned on language instructions. Figure~\ref{fig:architecture} gives an overview of LAGO's architecture.

\mypar{Vision Encoder.}
Given an RGB observation 
$o_t \in \mathbb{R}^{H \times W \times 3}$, the encoder produces a sequence of patch tokens:
\begin{equation}
z_t = \psi(o_t) \in \mathbb{R}^{N \times D_z}
\end{equation}
where $N$ is the number of patches and $D_z$ is the embedding dimension. LAGO is \emph{agnostic to the choice of vision encoder}, and can leverage either pretrained or jointly learned representations.

\mypar{Multi-Modal Conditioning.}
All conditioning signals are projected into a shared condition space of dimension $D_c$. A learned condition-type embedding $\tau \in \mathbb{R}^{D_c}$ is added to each conditioning vector to inform
the predictor of the operating mode.

\textit{Action conditioning.} For next-step transition prediction, low-level actions are encoded via a shared two-layer MLP $\phi_{act} : \mathbb{R}^{D_a} \to \mathbb{R}^{D_c}$. Continuous actions $a_t \in \mathbb{R}^{D_a}$ are passed to $\phi_{act}$ directly, while discrete actions are first one-hot encoded into $\mathbb{R}^{D_a}$ (with $D_a$ the number of discrete actions) and passed through the same MLP. The resulting conditioning vector is:
\begin{equation}
e_t^{act} = \phi_{act}(a_t), \qquad
c_t^{act} = e_t^{act} + \tau_{act} \in \mathbb{R}^{D_c}
\label{eq:act_cond_step_2}
\end{equation}

\textit{Subgoal conditioning.} For subgoal prediction, a language instruction is first embedded by
a language embedding model $\omega(\cdot)$ to obtain
$g \in \mathbb{R}^{D_L}$, then projected to $\mathbb{R}^{D_c}$ via a two-layer MLP
$\phi_{lang} : \mathbb{R}^{D_L} \to \mathbb{R}^{D_c}$. A completion scalar $\rho_t \in [0,1]$
estimating task progress is projected independently to $\mathbb{R}^{D_c}$ via a separate two-layer MLP
$\phi_{comp} : \mathbb{R} \to \mathbb{R}^{D_c}$. The subgoal conditioning vector combines both signals:
\begin{equation}
    e_t^{lang} = \phi_{lang}(g), \quad e_t^{comp} = \phi_{comp}(\rho_t)
    \label{eq:subgoal_cond_step_1}
\end{equation}
\begin{equation}
    c_t^{sg} = e_t^{lang} + e_t^{comp} + \tau_{sg} \in \mathbb{R}^{D_c}
    \label{eq:subgoal_cond_step_2}
\end{equation}

\textit{Conditioning injection.} The final condition vector $c_t \in \mathbb{R}^{D_c}$ (either
$c_t^{act}$ or $c_t^{sg}$ depending on the operating mode) is broadcast across all $N$ patch token
positions and concatenated with the latent $z_t \in \mathbb{R}^{N \times D_z}$, yielding the
conditioned input $z_t \oplus c_t \in \mathbb{R}^{N \times (D_z + D_c)}$.

\mypar{Latent Predictor.} The predictor $\pi(\cdot)$ operates over the conditioned token sequence $z_t \oplus c_t \in \mathbb{R}^{N \times (D_z + D_c)}$. Before returning the processed tokens, we remove the broadcasted condition tokens from the patch tokens to recover the original embedding dimension of the visual encoder $D_z$:
\begin{equation}
    \tilde{z} = \pi(z_t \oplus c_t) \in \mathbb{R}^{N \times D_z}
\end{equation}

\mypar{Training Objectives.}
The model is trained with a unified mean-squared error objective.

\textit{Next-step transition prediction.} Transition samples are drawn from an action-labeled interaction dataset. Each sample consists of a state observation $o_t$, the executed action $a_t$ and the subsequent observation $o_{t+1}$. Using $a_t$, we construct $c_t^{act}$ as specified in 
(\ref{eq:act_cond_step_2}), and train the predictor to minimize:
\begin{equation}
\mathcal{L}_{\text{act}} = \| \tilde{z}_{t+1} - z_{t+1} \|_2^2, \quad \tilde{z}_{t+1} = \pi(z_t \oplus c_t^{act})
\end{equation}

\textit{Subgoal prediction.} Subgoal samples are drawn from an action-less demonstration dataset. Given a demonstration episode of length $G$, we sample an initial state observation $o_j$ with index $j \in \{0, \dots, G-2\}$, and a subgoal state observation $o_k$ with index $k \in \{j+1, \dots, G-1\}$. From these, we set the completion scalar as
\begin{equation}
\rho = \frac{k - j}{G - 1 - j} \in (0,1].
\end{equation}
Intuitively, this encodes how far along the episode the subgoal lies relative to $o_j$. With this completion scalar, we construct $c^{sg}$ as described in (\ref{eq:subgoal_cond_step_1}) and (\ref{eq:subgoal_cond_step_2}), and train the predictor to minimize:
\begin{equation}
\mathcal{L}_{\text{subgoal}} = \| \tilde{z}_{\rho} - z_{\rho} \|_2^2, \quad \tilde{z}_{\rho} = \pi(z_j \oplus c^{sg}) 
\end{equation}

Because both tasks optimize the same latent-space prediction objective under different conditioning inputs, transition and subgoal samples can be freely intermixed within a single batch and optimized jointly. A single forward pass through $\pi$ thus learns both environment dynamics and subgoal structure, with no separate heads, losses, or training stages required.

\subsection{Planning with LAGO}

\begin{figure}[t!]
    \centering
    \includegraphics[width=.95\textwidth]{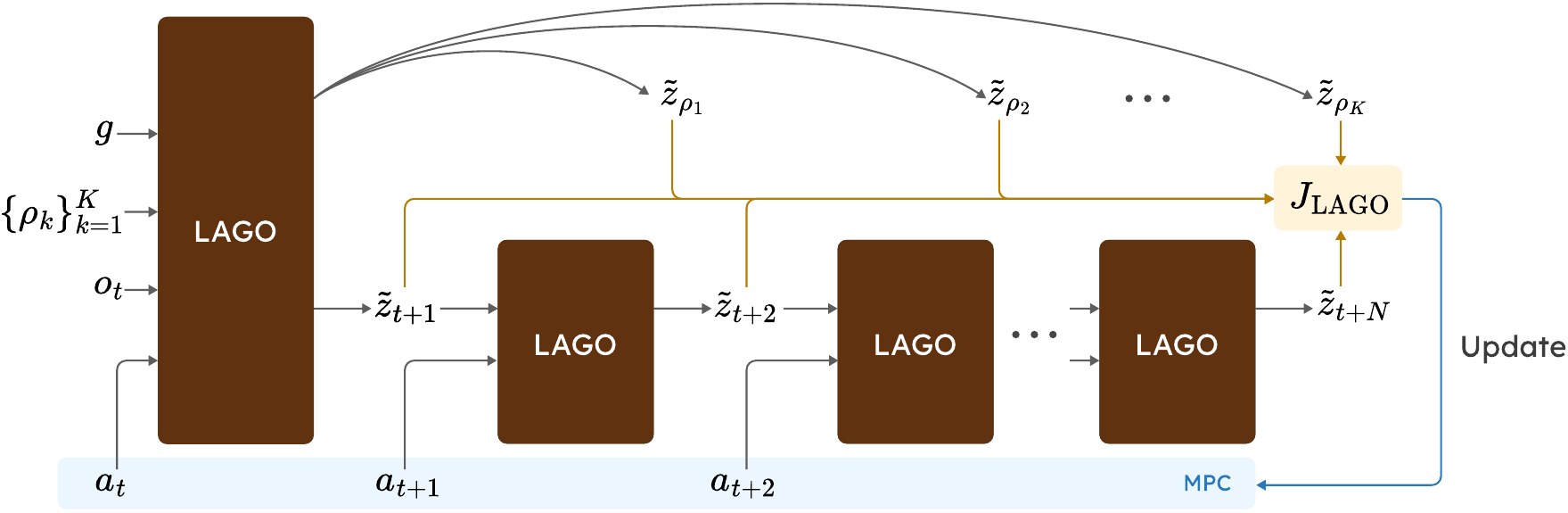} 
    \caption{\textbf{Planning with LAGO.} At each step, LAGO generates a sequence of $K$ latent subgoals from an arbitrary list of completion scalars $\{\rho_k\}_{k=1}^{K}$, thereby connecting the current state to the language-specified goal. A CEM planner then optimizes action sequences by evaluating candidate rollouts against the subgoals via $J_{\text{LAGO}}$, a soft minimum cost that measures the best alignment achieved between each subgoal and any timestep along the rollout. Subgoals are repredicted at every planning step, allowing the agent to continuously adapt to its evolving state.}
    \label{fig:planning}
    \vspace{-10pt}
\end{figure}

\mypar{Subgoal Generation.}
Given the current observation $o_t$ and a language-specified goal, we generate $K$ latent subgoals $\{z^{sg}_1, \dots, z^{sg}_K\}$ at uniformly spaced completion scalars $\rho_k = \frac{k}{K}$, $k \in \{1, \dots, K\}$ using the subgoal conditioning described in Section \ref{training_lago}. All subgoals are predicted independently from the current $z_t$ rather than autoregressively, avoiding compounding prediction errors. Subgoals are repredicted at every MPC step, so the sequence continuously adapts to the agent's evolving state.

\mypar{Action Optimization and Cost Formulation.}
At each planning step, the CEM solver samples candidate action sequences over a fixed horizon $H$, generating latent state rollouts $\tilde{z}_{1:H}$. To evaluate these rollouts against the predicted subgoals, we introduce a soft minimum trajectory cost that measures, for each subgoal, the best alignment achieved at any point along the rollout:
\begin{equation}
\label{eq:J-LAGO}
J_{\text{LAGO}} = \sum_{k=1}^{K} \lambda^{k} \!\cdot\! \left(\! -\tau \log \sum_{h=1}^{H} \exp\!\left(-\frac{1}{\tau}\text{MSE}\big(\tilde{z}_h,\, z^{sg}_k\big)\right) \right)
\end{equation}
where $\lambda\,{\in}\,(0, 1]$ is a subgoal decay factor and $\tau\,{>}\,0$ is a temperature parameter controlling the hardness of the temporal aggregation. For $\tau\,{\to}\,0$, the soft-min approaches a hard minimum, selecting only the closest timestep to each subgoal. As $\tau$ increases, timesteps near the minimum contribute proportionally more, favoring trajectories that converge to and remain near a subgoal over those that only transiently visit it. The decay $\lambda^k$ prioritizes earlier subgoals over distant ones. A visualization of the loss landscape produced by $J_{\text{LAGO}}$ can be found in Appendix~\ref{app:vis_loss_landscape}.

\vspace*{-5pt}
\section{Experiments}
\vspace*{-5pt}

\subsection{Experimental Setup}
\vspace*{-5pt}

\mypar{Implementation.} We use a frozen DINOv2 ViT-S/14~\citep{oquab2024dinov2learningrobustvisual} as our vision encoder 
$\psi$, which has been shown to provide a rich representation space for model-based planning \citep{zhou2025dinowmworldmodelspretrained}. 
Language instructions are embedded using EmbeddingGemma
300M~\citep{vera2025embeddinggemmapowerfullightweighttext}, yielding fixed-length representations for all language tasks. The latent predictor $\pi$ is implemented as a ViT~\citep{dosovitskiy2021imageworth16x16words}. To stress-test long-horizon planning, we train all methods with no frameskip and fix the planning horizon to 10 rollout steps. Full architectural details, 
hyperparameters, and optimization settings are provided in 
Appendix~\ref{hyperparameters}.

\mypar{Data Collection.} We rely on the SWM library~\citep{maes2026stableworldmodelv1reproducibleworldmodeling} for data collection in all environments except Kingdom. The suboptimal interaction dataset used to learn next-step transition dynamics is comprised of 50k environment transitions. While the majority of these transitions are collected using a random policy, we also include noisy expert policy rollouts. This mixture is necessary to provide coverage of both object manipulation in control tasks and room traversal in navigation tasks \citep{wang2026temporalstraighteninglatentplanning,maes2026leworldmodelstableendtoendjointembedding}. The demonstration dataset used to learn subgoal prediction is comprised of 10k language-labeled expert trajectories.

\mypar{Environments.} We evaluate on the following environments.
\textbf{(1)} TwoRoom~\citep{sobal2025learningrewardfreeofflinedata} is a 2D continuous-control navigation task with two rooms joined by a narrow doorway. The agent must navigate from a starting position to a target one.
\textbf{(2)} Kingdom is a 2D grid-world we introduce, with randomized layouts and irreversible water hazards that stress subgoal generation and long-horizon planning. The agent must navigate to a desired location. \textbf{(3)} OGBench Cube~\citep{park2025ogbenchbenchmarkingofflinegoalconditioned} is a manipulation task where a robotic arm moves a cube to a target location. We apply a distance-based gripper heuristic to all planners.
\textbf{(4)} OGBench Scene~\citep{park2025ogbenchbenchmarkingofflinegoalconditioned} is a fixed-layout manipulation task where the agent must interact with a drawer, window, and cube. While image goal baselines are provided an image of the agent in the desired location to solve the task, we adapt all environments to support language goals (e.g., ``place the red dot at position $[x,y]$'' or ``reach the left of the dragon''). Prior work \citep{zhou2025dinowmworldmodelspretrained, wang2026temporalstraighteninglatentplanning, maes2026leworldmodelstableendtoendjointembedding} evaluated methods in settings where targets are placed randomly within a fixed distance, often yielding near-solved or trivial instances (see Appendix~\ref{tworoom_evaluation_examples} for examples). In contrast, we set each level to a fixed start-to-target distance, eliminating the risk of such degenerate cases and stress-testing long-horizon planning. All environment-generation seeds used to construct the training datasets are stored with the corresponding data. Evaluation uses held-out environment seeds that do not overlap with any training seed. We report success rate, averaged over 10 seeds per difficulty level. Methods that plan with CEM are additionally repeated 5 times per seed, since their planning is stochastic. Full details can be found in Appendix~\ref{app_environments}.

\mypar{Image goal baselines.} We compare against the following recent world-model baselines: \textbf{(1)} DINO-WM \citep{zhou2025dinowmworldmodelspretrained} is a JEPA architecture using a frozen DINOv2~\citep{oquab2024dinov2learningrobustvisual} as its vision encoder. \textbf{(2)} Temporal Straightening \citep{wang2026temporalstraighteninglatentplanning} (T.S.) extends DINO-WM by adding a learned projection over the frozen patch tokens and a temporal straightening regularization term. This encourages the latent space to reduce curvature across successive states along expert trajectories. \textbf{(3)} LeWM \citep{maes2026leworldmodelstableendtoendjointembedding} is a fully end-to-end trained JEPA that learns both encoder and predictor from pixels, relying exclusively on the SIGReg regularizer \citep{balestriero2025lejepaprovablescalableselfsupervised} to prevent representation collapse. \textbf{(4)} HWM \citep{zhang2026hierarchicalplanninglatentworld} adds a high-level latent world model on top of a low-level one, operating over learned latent macro-actions at a coarser timescale. At test time, it plans macro-actions toward the goal image with the high-level model, and the first predicted latent state serves as a subgoal for low-level planning over primitive actions. \textbf{(5)} We introduce two oracle-subgoal (O.S.) variants, in which subgoals are replaced at every planning step by oracle ones, obtained by recomputing the shortest path to the goal and rendering equally spaced waypoints along it. LAGO~(O.S.) uses the same checkpoint as LAGO, but its predictor is used only for next-state prediction. DINO-WM~(O.S.) uses the same oracle subgoals and the same $J_{\text{LAGO}}$ planning objective, but obtains next-state predictions from the DINO-WM baseline.

To ensure a fair comparison with LAGO, all image-goal baselines are trained using next-step transition samples from both the interaction and the demonstration datasets. \textit{LAGO ignores the demonstration actions, using the action-less demonstrations solely for subgoal learning}. All methods are provided an equal compute budget and the evaluated checkpoints are selected based on validation accuracy. We provide visualizations of the decoded latent space for LeWM and Temporal Straightening in Appendix~\ref{latent_space_visualization}.

\mypar{Language Goal Baselines.}
We compare against the following language-conditioned baselines: \textbf{(1)} LIV \citep{ma2023livlanguageimagerepresentationsrewards} is a contrastive vision-language model trained on robotic interaction data with additional supervision for value estimation. We finetune LIV on the expert trajectories.
\textbf{(2)} PE-Core \citep{bolya2025perceptionencoderbestvisual} is a general-purpose vision-language encoder trained with a contrastive objective on large-scale multimodal data. It provides a strong off-the-shelf baseline for language grounding without robotics-specific adaptation. We use frozen representations without finetuning. \textbf{(3)} Robo-Dopamine \citep{tan2025robodopaminegeneralprocessreward} is a general-purpose reward modeling framework for robotic manipulation, built around a generative vision-language General Reward Model (GRM) that estimates task progress by conditioning on state images and language instructions. We use frozen representations without finetuning. \textbf{(4)} GCBC is a goal-conditioned behavior cloning policy trained to directly imitate the demonstration actions from the current state and language embedding, without a world model, subgoals, or any planning at test time.

LIV, PE-Core, and Robo-Dopamine are evaluated as reward models for MPC planning. Candidate rollouts are generated with the true environment dynamics and the resulting renders are scored directly by each method's vision-language reward.

\vspace*{-8pt}
\subsection{Performance of Planning with Self-Predicted Subgoals}
\label{sec:results-subgoal-performance}
\vspace*{-5pt}

\begin{figure}[t]
    \centering
    \includegraphics[width=\textwidth]{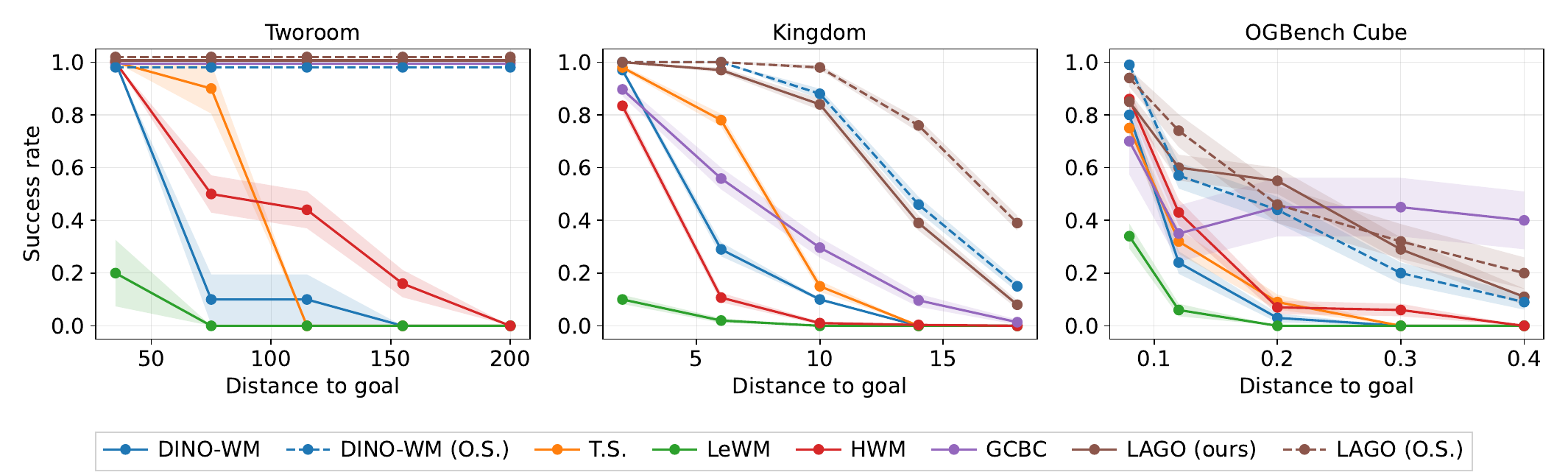} 
    \caption{\textbf{Performance of Planning with Self-Predicted Subgoals.} Success rate vs. distance-to-goal on TwoRoom, Kingdom (averaged over all goals), and OGBench Cube. (O.S.): oracle subgoals.}
    \label{fig:main_results}
    \vspace{-12pt}
\end{figure}

We evaluate LAGO against image-goal and language-conditioned baselines on three environments of increasing planning difficulty, asking: \textbf{(i)} ``\emph{does LAGO's subgoal prediction improve long-horizon planning over baselines?}'' \textbf{(ii)} ``\emph{how much is lost because subgoals are predicted rather than perfect?}'' \textbf{(iii)} ``\emph{does the dual-objective design help beyond the subgoals themselves?}'' To answer (ii) and (iii), we use the (O.S.) variants. LAGO and DINO-WM use the same frozen DINOv2 representation space and predictor architecture, making DINO-WM~(O.S.) a controlled comparison in which the subgoal branch is replaced by oracle subgoals and the two prediction modes are trained separately. Thus, comparing LAGO with LAGO~(O.S.) measures the performance cost associated with predicting subgoals, while comparing LAGO~(O.S.) with DINO-WM~(O.S.) tests whether the joint training of subgoal and transition prediction improves the learned planning dynamics. Figure \ref{fig:main_results} reports success rate as a function of environment difficulty; Kingdom is averaged over all possible goals.

LAGO matches GCBC and both oracle-subgoal variants on TwoRoom, where they all achieve 100\% success at every difficulty. It substantially exceeds
Temporal Straightening (T.S.) and LeWM at the three longest distances, where
both methods reach zero success, while HWM degrades more gradually but also
reaches zero. Kingdom provides the clearest long-horizon test as shortest paths
are often highly curved around obstacles, and successful trajectories must
anticipate terminal hazards such as water tiles before reaching them. LAGO outperforms every non-oracle baseline at every difficulty on Kingdom, while GCBC
degrades from 90\% to 1\% success as difficulty increases. At difficulty 10,
for example, LAGO achieves 84\% success compared with 30\% for GCBC. This
gap is consistent with the advantage of explicitly planning through
intermediate states rather than directly predicting actions from the current
observation and language goal. On OGBench Cube, the gap between methods is smaller because the task mainly requires reaching the cube and then moving it to the target, with little of the detour planning and terminal-hazard avoidance that distinguishes Kingdom. LAGO nevertheless outperforms GCBC at the two lowest difficulties and is comparable at 0.2, while GCBC is stronger at the two largest (45\% vs 29\% and 40\% vs 11\%), likely because grasp dynamics are hard for action-conditioned rollouts to model.

The oracle-subgoal comparisons locate LAGO's remaining performance gap. On
TwoRoom and OGBench Cube, LAGO matches LAGO~(O.S.), indicating that predicted subgoals are sufficiently accurate on these tasks. On Kingdom, LAGO performs comparably to LAGO (O.S.) at difficulties 2 and 6, but the gap widens from difficulty 10 onward. This suggests that subgoal quality, rather than the planning procedure, is the primary bottleneck at the longest horizons. Separately, LAGO (O.S.) exceeds DINO-WM (O.S.) in most unsaturated settings, most clearly at long Kingdom horizons (76\% vs 46\% at difficulty 14 and 39\% vs 15\% at difficulty 18). This supports the hypothesis that joint \emph{dual-objective training improves the transition model used for planning}, and the benefit is not explained by subgoal prediction alone. On OGBench Cube, LAGO with predicted subgoals matches or exceeds DINO-WM~(O.S.) at the four longest difficulties, showing that the jointly trained model can compensate for imperfect subgoals.

\vspace*{-10pt}
\subsection{Language Grounding Performance}
\label{sec:language-grounding}
\vspace*{-5pt}

\begin{figure}[t]
    \centering
    \includegraphics[width=0.9\textwidth]{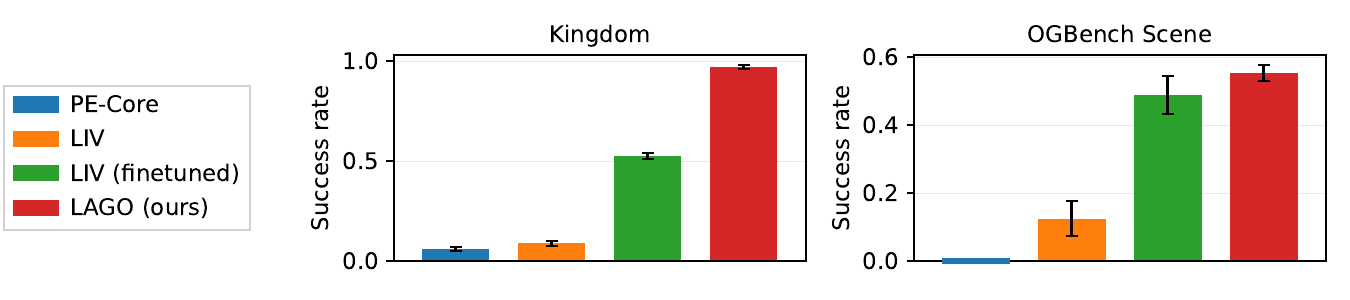}
    
    \vspace{-10pt}    
    \caption{\textbf{LAGO vs. vision-language reward baselines.} Success rate averaged across all language goals on Kingdom (difficulty 6) and OGBench Scene. Baselines score rollouts simulated with true dynamics; LAGO plans in latent space.}
    \label{fig:all_envs_language}
    \vspace{-8pt}
\end{figure}

We evaluate the performance of grounding language directly in the world model's representation space, as opposed to offloading progress estimation to a dedicated vision-language model. We compare LAGO against language-goal baselines evaluated on Kingdom and OGBench Scene across all available tasks at fixed difficulty: Kingdom is set to 6 steps, and OGBench Scene fixes the starting position to the center of the environment. Results are reported as success rates in Figure~\ref{fig:all_envs_language}, showing that PE-Core provides no meaningful signal for language-conditioned planning (6\% in Kingdom, 0\% in OGBench Scene), and LIV exhibits a similar trend (8.8\% and 12.5\%, respectively). LIV-finetuned is stronger due to its direct adaptation to each environment's expert demonstrations, yet still falls short of LAGO ($+44.4$\% in Kingdom and $+6.6$\% in OGBench Scene), indicating that grounding language within the world model's representation space outperforms delegating this grounding to a separate, dedicated VLM. We further study the effect of scaling language understanding independently from next-state prediction in Appendix~\ref{app:LAGOSET}.

Robo-Dopamine \citep{tan2025robodopaminegeneralprocessreward} is a large generative reward model, poorly suited to the high-sampling demands of model-based planning. Over 200 GPU-synced trials at batch size 1, it takes $451 \pm 13$ms per observation, versus $3.7 \pm 1.4$ms for LIV-finetuned and $11.0 \pm 0.7$ms for LAGO ($K\,{=}\,5$). This 121$\times$ slowdown rules it out as a reward signal for MPC under practical time constraints, so we instead assess its reward quality with the VOC metric \citep{ma2024visionlanguagemodelsincontext} in Appendix~\ref{sec:robodopamine-voc}.

\vspace*{-5pt}
\subsection{Model Analysis}
\vspace*{-5pt}

For this analysis, we fix difficulty at 155 for TwoRoom and 10 for Kingdom. Complete results tables can be found in Appendix~\ref{app:complete_results_table}.

\begin{figure}[t]
    \centering
    \includegraphics[width=0.9\textwidth]{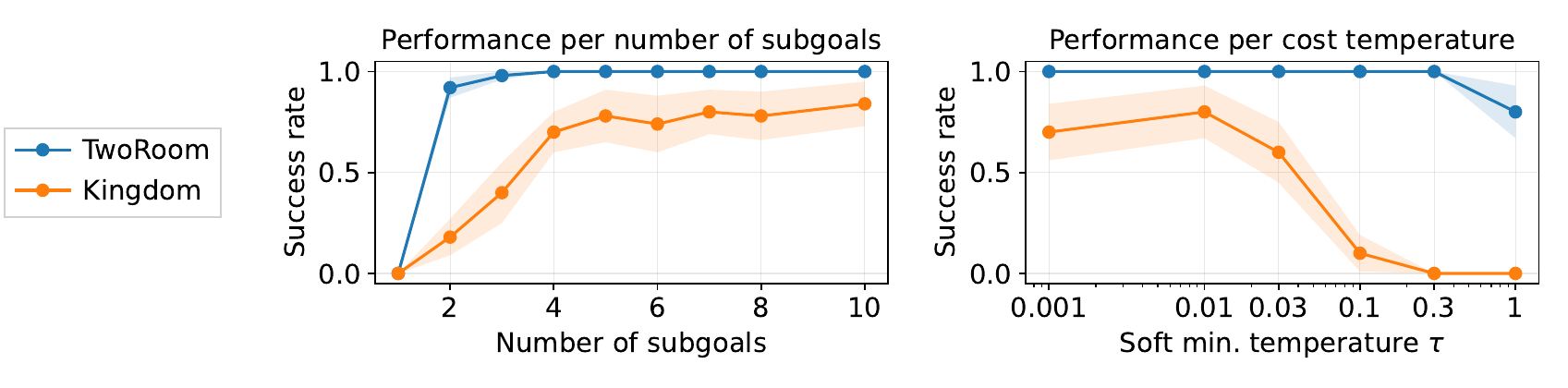}

    \vspace{-10pt}    
    \caption{\textbf{Ablations of LAGO's planning components.} Success rates on Kingdom and TwoRoom as a function of
    \textbf{(left)} the number of subgoals $K$ and
    \textbf{(right)} the cost temperature $\tau$.}
    \label{fig:ablation_joint}
    \vspace{-10pt}
\end{figure}

\mypar{Number of Subgoals.} We vary $K \in \{1, \dots, 10\}$ with all other hyperparameters fixed (Figure~\ref{fig:ablation_joint}). A single subgoal fails in both environments, and performance is poor at low $K$ because the planner must bridge large latent distances between subgoals. TwoRoom recovers almost immediately, while Kingdom recovers gradually. From $K\,{=}\,5$ onward, performance plateaus (TwoRoom at 100\%, Kingdom within error), so LAGO is insensitive to $K$ once moderately large. We use $K\,{=}\,5$ throughout our experiments.

\mypar{Cost Function Analysis.} We sweep the soft-min temperature $\tau$ of $J_{\text{LAGO}}$ (Eq.~\ref{eq:J-LAGO}), which controls how broadly each subgoal is matched across rollout timesteps (Figure~\ref{fig:ablation_joint}). Low $\tau$ focuses on the best-aligned timestep, so each subgoal need only be visited at some point; high $\tau$ approaches a uniform sum that pulls every step toward every subgoal, implicitly enforcing straight-line interpolation. Lower temperatures perform substantially better: Kingdom peaks at $80\%$ for $\tau\,{=}\,0.01$ and degrades monotonically as temperature increases, falling to $0\%$ at $\tau\,{=}\,0.3$. Meanwhile, TwoRoom remains saturated ($100\%$) across $\tau \le 0.3$ and drops only mildly to $80\%$ at $\tau\,{=}\,1.0$. This is consistent with Kingdom's shortest paths curving around obstacles, which requires rollout flexibility to avoid terminal water tiles. We select $\tau\,{=}\,0.001$ throughout.

\vspace*{-8pt}
\section{Conclusion}
\vspace*{-6pt}

We introduced LAGO, a hierarchical world model in which a single latent predictor forecasts both action-conditioned dynamics and language-conditioned subgoals. Planning against these subgoals with a soft minimum alignment cost turns a long-horizon instruction into a sequence of locally tractable objectives. The gains are largest where optimal paths are non-trivial. On Kingdom, whose shortest paths curve around terminal hazards, LAGO reaches 84\% success against 30\% for GCBC. Grounding language in the world model's own latent space also outperforms dedicated vision-language reward models with access to ground-truth dynamics.


\section*{AI use statement}

In this work, we used generative AI tools to implement methods, edit a research paper to improve readability, sourcing/searching for information, identify relevant literature.
We have not used generative AI tools to generate synthetic data sets, help develop theoretical models or conceptual frameworks, formulate mathematical claims, provide critical ingredients for proving mathematical claims, assist in the writing of proofs, propose or refine hypotheses, design or provide feedback on research  methodology or experiments, assist with translation, clean and reformat dataset, support qualitative and thematic data analysis, interpret results. We have reviewed all AI-assisted work. All experiments were hand reviewed and tested by two of the authors to ensure correctness. We take responsibility for the final content of this work,
including text, claims or artifacts produced with the aid of generative AI.

\bibliography{iclr2027_conference}

@inproceedings{hafner2020dream,
  title={Dream to Control: Learning Behaviors by Latent Imagination},
  author={Hafner, Danijar and Lillicrap, Timothy and Ba, Jimmy and Norouzi, Mohammad},
  booktitle={ICLR},
  year={2020}
}

@inproceedings{huang2024subgoal,
  title={Subgoal Diffuser: Coarse-to-fine subgoal generation to guide model predictive control for robot manipulation},
  author={Huang, Zixuan and Lin, Yating and Yang, Fan and Berenson, Dmitry},
  booktitle={2024 IEEE International Conference on Robotics and Automation (ICRA)},
  pages={16489--16495},
  year={2024},
  organization={IEEE}
}

@misc{ma2024visionlanguagemodelsincontext,
      title={Vision Language Models are In-Context Value Learners}, 
      author={Yecheng Jason Ma and Joey Hejna and Ayzaan Wahid and Chuyuan Fu and Dhruv Shah and Jacky Liang and Zhuo Xu and Sean Kirmani and Peng Xu and Danny Driess and Ted Xiao and Jonathan Tompson and Osbert Bastani and Dinesh Jayaraman and Wenhao Yu and Tingnan Zhang and Dorsa Sadigh and Fei Xia},
      year={2024},
      eprint={2411.04549},
      archivePrefix={arXiv},
      primaryClass={cs.RO},
      url={https://arxiv.org/abs/2411.04549}, 
}

@article{haramati2026hierarchical,
  title={Hierarchical Entity-centric Reinforcement Learning with Factored Subgoal Diffusion},
  author={Haramati, Dan and Qi, Carl and Daniel, Tal and Zhang, Amy and Tamar, Aviv and Konidaris, George},
  journal={arXiv preprint arXiv:2602.02722},
  year={2026}
}

@misc{zhou2025dinowmworldmodelspretrained,
      title={DINO-WM: World Models on Pre-trained Visual Features enable Zero-shot Planning}, 
      author={Gaoyue Zhou and Hengkai Pan and Yann LeCun and Lerrel Pinto},
      year={2025},
      eprint={2411.04983},
      archivePrefix={arXiv},
      primaryClass={cs.RO},
      url={https://arxiv.org/abs/2411.04983}, 
}

@misc{assran2023selfsupervisedlearningimagesjointembedding,
      title={Self-Supervised Learning from Images with a Joint-Embedding Predictive Architecture}, 
      author={Mahmoud Assran and Quentin Duval and Ishan Misra and Piotr Bojanowski and Pascal Vincent and Michael Rabbat and Yann LeCun and Nicolas Ballas},
      year={2023},
      eprint={2301.08243},
      archivePrefix={arXiv},
      primaryClass={cs.CV},
      url={https://arxiv.org/abs/2301.08243}, 
}

@misc{maes2026leworldmodelstableendtoendjointembedding,
      title={LeWorldModel: Stable End-to-End Joint-Embedding Predictive Architecture from Pixels}, 
      author={Lucas Maes and Quentin Le Lidec and Damien Scieur and Yann LeCun and Randall Balestriero},
      year={2026},
      eprint={2603.19312},
      archivePrefix={arXiv},
      primaryClass={cs.LG},
      url={https://arxiv.org/abs/2603.19312}, 
}

@misc{oquab2024dinov2learningrobustvisual,
      title={DINOv2: Learning Robust Visual Features without Supervision}, 
      author={Maxime Oquab and Timothée Darcet and Théo Moutakanni and Huy Vo and Marc Szafraniec and Vasil Khalidov and Pierre Fernandez and Daniel Haziza and Francisco Massa and Alaaeldin El-Nouby and Mahmoud Assran and Nicolas Ballas and Wojciech Galuba and Russell Howes and Po-Yao Huang and Shang-Wen Li and Ishan Misra and Michael Rabbat and Vasu Sharma and Gabriel Synnaeve and Hu Xu and Hervé Jegou and Julien Mairal and Patrick Labatut and Armand Joulin and Piotr Bojanowski},
      year={2024},
      eprint={2304.07193},
      archivePrefix={arXiv},
      primaryClass={cs.CV},
      url={https://arxiv.org/abs/2304.07193}, 
}

@misc{vera2025embeddinggemmapowerfullightweighttext,
      title={EmbeddingGemma: Powerful and Lightweight Text Representations}, 
      author={Henrique Schechter Vera and Sahil Dua and Biao Zhang and Daniel Salz and Ryan Mullins and Sindhu Raghuram Panyam and Sara Smoot and Iftekhar Naim and Joe Zou and Feiyang Chen and Daniel Cer and Alice Lisak and Min Choi and Lucas Gonzalez and Omar Sanseviero and Glenn Cameron and Ian Ballantyne and Kat Black and Kaifeng Chen and Weiyi Wang and Zhe Li and Gus Martins and Jinhyuk Lee and Mark Sherwood and Juyeong Ji and Renjie Wu and Jingxiao Zheng and Jyotinder Singh and Abheesht Sharma and Divyashree Sreepathihalli and Aashi Jain and Adham Elarabawy and AJ Co and Andreas Doumanoglou and Babak Samari and Ben Hora and Brian Potetz and Dahun Kim and Enrique Alfonseca and Fedor Moiseev and Feng Han and Frank Palma Gomez and Gustavo Hernández Ábrego and Hesen Zhang and Hui Hui and Jay Han and Karan Gill and Ke Chen and Koert Chen and Madhuri Shanbhogue and Michael Boratko and Paul Suganthan and Sai Meher Karthik Duddu and Sandeep Mariserla and Setareh Ariafar and Shanfeng Zhang and Shijie Zhang and Simon Baumgartner and Sonam Goenka and Steve Qiu and Tanmaya Dabral and Trevor Walker and Vikram Rao and Waleed Khawaja and Wenlei Zhou and Xiaoqi Ren and Ye Xia and Yichang Chen and Yi-Ting Chen and Zhe Dong and Zhongli Ding and Francesco Visin and Gaël Liu and Jiageng Zhang and Kathleen Kenealy and Michelle Casbon and Ravin Kumar and Thomas Mesnard and Zach Gleicher and Cormac Brick and Olivier Lacombe and Adam Roberts and Qin Yin and Yunhsuan Sung and Raphael Hoffmann and Tris Warkentin and Armand Joulin and Tom Duerig and Mojtaba Seyedhosseini},
      year={2025},
      eprint={2509.20354},
      archivePrefix={arXiv},
      primaryClass={cs.CL},
      url={https://arxiv.org/abs/2509.20354}, 
}

@misc{dosovitskiy2021imageworth16x16words,
      title={An Image is Worth 16x16 Words: Transformers for Image Recognition at Scale}, 
      author={Alexey Dosovitskiy and Lucas Beyer and Alexander Kolesnikov and Dirk Weissenborn and Xiaohua Zhai and Thomas Unterthiner and Mostafa Dehghani and Matthias Minderer and Georg Heigold and Sylvain Gelly and Jakob Uszkoreit and Neil Houlsby},
      year={2021},
      eprint={2010.11929},
      archivePrefix={arXiv},
      primaryClass={cs.CV},
      url={https://arxiv.org/abs/2010.11929}, 
}

@inproceedings{hafner2019planet,
  title     = {Learning Latent Dynamics for Planning from Pixels},
  author    = {Hafner, Danijar and Lillicrap, Timothy and Ba, Jimmy and Norouzi, Mohammad},
  booktitle = {International Conference on Machine Learning (ICML)},
  year      = {2019}
}

@misc{nair2022r3muniversalvisualrepresentation,
      title={R3M: A Universal Visual Representation for Robot Manipulation}, 
      author={Suraj Nair and Aravind Rajeswaran and Vikash Kumar and Chelsea Finn and Abhinav Gupta},
      year={2022},
      eprint={2203.12601},
      archivePrefix={arXiv},
      primaryClass={cs.RO},
      url={https://arxiv.org/abs/2203.12601}, 
}

@article{hafner2025mastering,
  title={Mastering diverse control tasks through world models},
  author={Hafner, Danijar and Pasukonis, Jurgis and Ba, Jimmy and Lillicrap, Timothy},
  journal={Nature},
  volume={640},
  number={8059},
  pages={647--653},
  year={2025},
  publisher={Nature Publishing Group UK London}
}

@inproceedings{hansen2022tdmpc,
  title     = {Temporal Difference Learning for Model Predictive Control},
  author    = {Hansen, Nicklas and Su, Hao and Wang, Xiaolong},
  booktitle = {International Conference on Machine Learning (ICML)},
  year      = {2022}
}

@inproceedings{chen2024simple,
title={Simple Hierarchical Planning with Diffusion},
author={Chang Chen and Fei Deng and Kenji Kawaguchi and Caglar Gulcehre and Sungjin Ahn},
booktitle={The Twelfth International Conference on Learning Representations},
year={2024},
url={https://openreview.net/forum?id=kXHEBK9uAY}
}

@inproceedings{wang2024rlvlmf,
  title     = {RL-VLM-F: Reinforcement Learning with Vision-Language Foundation Models},
  author    = {Wang, Zihan and others},
  booktitle = {International Conference on Machine Learning (ICML)},
  year      = {2024}
}

@misc{roy2025revisitinglearningobjectivesvisionlanguage,
      title={Revisiting the Learning Objectives of Vision-Language Reward Models}, 
      author={Simon Roy and Samuel Barbeau and Giovanni Beltrame and Christian Desrosiers and Nicolas Thome},
      year={2025},
      eprint={2512.20675},
      archivePrefix={arXiv},
      primaryClass={cs.LG},
      url={https://arxiv.org/abs/2512.20675}, 
}

@article{https://doi.org/10.5281/zenodo.1207631,
  doi = {10.5281/ZENODO.1207631},
  url = {https://zenodo.org/record/1207631},
  author = {Ha, David and Schmidhuber, Jürgen},
  title = {World Models},
  publisher = {Zenodo},
  year = {2018},
  journal = {Zenodo},
  copyright = {Creative Commons Attribution 4.0}
}

@misc{rocamonde2024visionlanguagemodelszeroshotreward,
      title={Vision-Language Models are Zero-Shot Reward Models for Reinforcement Learning}, 
      author={Juan Rocamonde and Victoriano Montesinos and Elvis Nava and Ethan Perez and David Lindner},
      year={2024},
      eprint={2310.12921},
      archivePrefix={arXiv},
      primaryClass={cs.LG},
      url={https://arxiv.org/abs/2310.12921}, 
}

@misc{assran2025vjepa2selfsupervisedvideo,
      title={V-JEPA 2: Self-Supervised Video Models Enable Understanding, Prediction and Planning}, 
      author={Mido Assran and Adrien Bardes and David Fan and Quentin Garrido and Russell Howes and Mojtaba and Komeili and Matthew Muckley and Ammar Rizvi and Claire Roberts and Koustuv Sinha and Artem Zholus and Sergio Arnaud and Abha Gejji and Ada Martin and Francois Robert Hogan and Daniel Dugas and Piotr Bojanowski and Vasil Khalidov and Patrick Labatut and Francisco Massa and Marc Szafraniec and Kapil Krishnakumar and Yong Li and Xiaodong Ma and Sarath Chandar and Franziska Meier and Yann LeCun and Michael Rabbat and Nicolas Ballas},
      year={2025},
      eprint={2506.09985},
      archivePrefix={arXiv},
      primaryClass={cs.AI},
      url={https://arxiv.org/abs/2506.09985}, 
}

@misc{daniel2026latentparticleworldmodels,
      title={Latent Particle World Models: Self-supervised Object-centric Stochastic Dynamics Modeling}, 
      author={Tal Daniel and Carl Qi and Dan Haramati and Amir Zadeh and Chuan Li and Aviv Tamar and Deepak Pathak and David Held},
      year={2026},
      eprint={2603.04553},
      archivePrefix={arXiv},
      primaryClass={cs.LG},
      url={https://arxiv.org/abs/2603.04553}, 
}

@inproceedings{brohan2023rt2,
  title     = {RT-2: Vision-Language-Action Models Transfer Web Knowledge to Robotic Control},
  author    = {Brohan, Anthony and others},
  booktitle = {Conference on Robot Learning (CoRL)},
  year      = {2023}
}

@misc{sobal2025learningrewardfreeofflinedata,
      title={Learning from Reward-Free Offline Data: A Case for Planning with Latent Dynamics Models}, 
      author={Vlad Sobal and Wancong Zhang and Kyunghyun Cho and Randall Balestriero and Tim G. J. Rudner and Yann LeCun},
      year={2025},
      eprint={2502.14819},
      archivePrefix={arXiv},
      primaryClass={cs.LG},
      url={https://arxiv.org/abs/2502.14819}, 
}

@misc{maes2026stableworldmodelv1reproducibleworldmodeling,
      title={stable-worldmodel-v1: Reproducible World Modeling Research and Evaluation}, 
      author={Lucas Maes and Quentin Le Lidec and Dan Haramati and Nassim Massaudi and Damien Scieur and Yann LeCun and Randall Balestriero},
      year={2026},
      eprint={2602.08968},
      archivePrefix={arXiv},
      primaryClass={cs.AI},
      url={https://arxiv.org/abs/2602.08968}, 
}

@misc{park2025ogbenchbenchmarkingofflinegoalconditioned,
      title={OGBench: Benchmarking Offline Goal-Conditioned RL}, 
      author={Seohong Park and Kevin Frans and Benjamin Eysenbach and Sergey Levine},
      year={2025},
      eprint={2410.20092},
      archivePrefix={arXiv},
      primaryClass={cs.LG},
      url={https://arxiv.org/abs/2410.20092}, 
}

@misc{wang2026temporalstraighteninglatentplanning,
      title={Temporal Straightening for Latent Planning}, 
      author={Ying Wang and Oumayma Bounou and Gaoyue Zhou and Randall Balestriero and Tim G. J. Rudner and Yann LeCun and Mengye Ren},
      year={2026},
      eprint={2603.12231},
      archivePrefix={arXiv},
      primaryClass={cs.LG},
      url={https://arxiv.org/abs/2603.12231}, 
}

@misc{balestriero2025lejepaprovablescalableselfsupervised,
      title={LeJEPA: Provable and Scalable Self-Supervised Learning Without the Heuristics}, 
      author={Randall Balestriero and Yann LeCun},
      year={2025},
      eprint={2511.08544},
      archivePrefix={arXiv},
      primaryClass={cs.LG},
      url={https://arxiv.org/abs/2511.08544}, 
}

@misc{ma2023livlanguageimagerepresentationsrewards,
      title={LIV: Language-Image Representations and Rewards for Robotic Control}, 
      author={Yecheng Jason Ma and William Liang and Vaidehi Som and Vikash Kumar and Amy Zhang and Osbert Bastani and Dinesh Jayaraman},
      year={2023},
      eprint={2306.00958},
      archivePrefix={arXiv},
      primaryClass={cs.RO},
      url={https://arxiv.org/abs/2306.00958}, 
}

@misc{bolya2025perceptionencoderbestvisual,
      title={Perception Encoder: The best visual embeddings are not at the output of the network}, 
      author={Daniel Bolya and Po-Yao Huang and Peize Sun and Jang Hyun Cho and Andrea Madotto and Chen Wei and Tengyu Ma and Jiale Zhi and Jathushan Rajasegaran and Hanoona Rasheed and Junke Wang and Marco Monteiro and Hu Xu and Shiyu Dong and Nikhila Ravi and Daniel Li and Piotr Dollár and Christoph Feichtenhofer},
      year={2025},
      eprint={2504.13181},
      archivePrefix={arXiv},
      primaryClass={cs.CV},
      url={https://arxiv.org/abs/2504.13181}, 
}

@misc{tan2025robodopaminegeneralprocessreward,
      title={Robo-Dopamine: General Process Reward Modeling for High-Precision Robotic Manipulation}, 
      author={Huajie Tan and Sixiang Chen and Yijie Xu and Zixiao Wang and Yuheng Ji and Cheng Chi and Yaoxu Lyu and Zhongxia Zhao and Xiansheng Chen and Peterson Co and Shaoxuan Xie and Guocai Yao and Pengwei Wang and Zhongyuan Wang and Shanghang Zhang},
      year={2025},
      eprint={2512.23703},
      archivePrefix={arXiv},
      primaryClass={cs.RO},
      url={https://arxiv.org/abs/2512.23703}, 
}

@inproceedings{hao2025chd,
  title={CHD: Coupled Hierarchical Diffusion for Long-Horizon Tasks},
  author={Hao, Ce and Xiao, Anxing and Xue, Zhiwei and Soh, Harold},
  booktitle={Conference on Robot Learning},
  pages={1420--1451},
  year={2025},
  organization={PMLR}
}

@book{sutton2018reinforcement,
  title={Reinforcement learning: An introduction},
  author={Sutton, Richard S and Barto, Andrew G},
  year={2018},
  edition={2nd},
  publisher={The MIT Press},
  address={Cambridge, MA}
}

@misc{hansen2025learningmassivelymultitaskworld,
      title={Learning Massively Multitask World Models for Continuous Control}, 
      author={Nicklas Hansen and Hao Su and Xiaolong Wang},
      year={2025},
      eprint={2511.19584},
      archivePrefix={arXiv},
      primaryClass={cs.LG},
      url={https://arxiv.org/abs/2511.19584}, 
}

@misc{nair2021learninglanguageconditionedrobotbehavior,
      title={Learning Language-Conditioned Robot Behavior from Offline Data and Crowd-Sourced Annotation}, 
      author={Suraj Nair and Eric Mitchell and Kevin Chen and Brian Ichter and Silvio Savarese and Chelsea Finn},
      year={2021},
      eprint={2109.01115},
      archivePrefix={arXiv},
      primaryClass={cs.RO},
      url={https://arxiv.org/abs/2109.01115}, 
}

@misc{sontakke2023roboclipdemonstrationlearnrobot,
      title={RoboCLIP: One Demonstration is Enough to Learn Robot Policies}, 
      author={Sumedh A Sontakke and Jesse Zhang and Sébastien M. R. Arnold and Karl Pertsch and Erdem Bıyık and Dorsa Sadigh and Chelsea Finn and Laurent Itti},
      year={2023},
      eprint={2310.07899},
      archivePrefix={arXiv},
      primaryClass={cs.AI},
      url={https://arxiv.org/abs/2310.07899}, 
}

@misc{intelligence2025pi05visionlanguageactionmodelopenworld,
      title={$\pi_{0.5}$: a Vision-Language-Action Model with Open-World Generalization}, 
      author={Kevin Black and Noah Brown and James Darpinian and Karan Dhabalia and Danny Driess and Adnan Esmail and Michael Equi and Chelsea Finn and Niccolo Fusai and Manuel Y. Galliker and Dibya Ghosh and Lachy Groom and Karol Hausman and Brian Ichter and Szymon Jakubczak and Tim Jones and Liyiming Ke and Devin LeBlanc and Sergey Levine and Adrian Li-Bell and Mohith Mothukuri and Suraj Nair and Karl Pertsch and Allen Z. Ren and Lucy Xiaoyang Shi and Laura Smith and Jost Tobias Springenberg and Kyle Stachowicz and James Tanner and Quan Vuong and Homer Walke and Anna Walling and Haohuan Wang and Lili Yu and Ury Zhilinsky},
      year={2025},
      eprint={2504.16054},
      archivePrefix={arXiv},
      primaryClass={cs.LG},
      url={https://arxiv.org/abs/2504.16054}, 
}

@misc{fu2024furlvisuallanguagemodelsfuzzy,
      title={FuRL: Visual-Language Models as Fuzzy Rewards for Reinforcement Learning}, 
      author={Yuwei Fu and Haichao Zhang and Di Wu and Wei Xu and Benoit Boulet},
      year={2024},
      eprint={2406.00645},
      archivePrefix={arXiv},
      primaryClass={cs.LG},
      url={https://arxiv.org/abs/2406.00645}, 
}

@misc{du2023learninguniversalpoliciestextguided,
      title={Learning Universal Policies via Text-Guided Video Generation}, 
      author={Yilun Du and Mengjiao Yang and Bo Dai and Hanjun Dai and Ofir Nachum and Joshua B. Tenenbaum and Dale Schuurmans and Pieter Abbeel},
      year={2023},
      eprint={2302.00111},
      archivePrefix={arXiv},
      primaryClass={cs.AI},
      url={https://arxiv.org/abs/2302.00111}, 
}

@misc{black2023zeroshotroboticmanipulationpretrained,
      title={Zero-Shot Robotic Manipulation with Pretrained Image-Editing Diffusion Models}, 
      author={Kevin Black and Mitsuhiko Nakamoto and Pranav Atreya and Homer Walke and Chelsea Finn and Aviral Kumar and Sergey Levine},
      year={2023},
      eprint={2310.10639},
      archivePrefix={arXiv},
      primaryClass={cs.RO},
      url={https://arxiv.org/abs/2310.10639}, 
}

@misc{ajay2023compositionalfoundationmodelshierarchical,
      title={Compositional Foundation Models for Hierarchical Planning}, 
      author={Anurag Ajay and Seungwook Han and Yilun Du and Shuang Li and Abhi Gupta and Tommi Jaakkola and Josh Tenenbaum and Leslie Kaelbling and Akash Srivastava and Pulkit Agrawal},
      year={2023},
      eprint={2309.08587},
      archivePrefix={arXiv},
      primaryClass={cs.LG},
      url={https://arxiv.org/abs/2309.08587}, 
}

@misc{ko2023learningactactionlessvideos,
      title={Learning to Act from Actionless Videos through Dense Correspondences}, 
      author={Po-Chen Ko and Jiayuan Mao and Yilun Du and Shao-Hua Sun and Joshua B. Tenenbaum},
      year={2023},
      eprint={2310.08576},
      archivePrefix={arXiv},
      primaryClass={cs.RO},
      url={https://arxiv.org/abs/2310.08576}, 
}

@misc{du2023videolanguageplanning,
      title={Video Language Planning}, 
      author={Yilun Du and Mengjiao Yang and Pete Florence and Fei Xia and Ayzaan Wahid and Brian Ichter and Pierre Sermanet and Tianhe Yu and Pieter Abbeel and Joshua B. Tenenbaum and Leslie Kaelbling and Andy Zeng and Jonathan Tompson},
      year={2023},
      eprint={2310.10625},
      archivePrefix={arXiv},
      primaryClass={cs.CV},
      url={https://arxiv.org/abs/2310.10625}, 
}

@misc{grauman2022ego4dworld3000hours,
      title={Ego4D: Around the World in 3,000 Hours of Egocentric Video}, 
      author={Kristen Grauman and Andrew Westbury and Eugene Byrne and Zachary Chavis and Antonino Furnari and Rohit Girdhar and Jackson Hamburger and Hao Jiang and Miao Liu and Xingyu Liu and Miguel Martin and Tushar Nagarajan and Ilija Radosavovic and Santhosh Kumar Ramakrishnan and Fiona Ryan and Jayant Sharma and Michael Wray and Mengmeng Xu and Eric Zhongcong Xu and Chen Zhao and Siddhant Bansal and Dhruv Batra and Vincent Cartillier and Sean Crane and Tien Do and Morrie Doulaty and Akshay Erapalli and Christoph Feichtenhofer and Adriano Fragomeni and Qichen Fu and Abrham Gebreselasie and Cristina Gonzalez and James Hillis and Xuhua Huang and Yifei Huang and Wenqi Jia and Weslie Khoo and Jachym Kolar and Satwik Kottur and Anurag Kumar and Federico Landini and Chao Li and Yanghao Li and Zhenqiang Li and Karttikeya Mangalam and Raghava Modhugu and Jonathan Munro and Tullie Murrell and Takumi Nishiyasu and Will Price and Paola Ruiz Puentes and Merey Ramazanova and Leda Sari and Kiran Somasundaram and Audrey Southerland and Yusuke Sugano and Ruijie Tao and Minh Vo and Yuchen Wang and Xindi Wu and Takuma Yagi and Ziwei Zhao and Yunyi Zhu and Pablo Arbelaez and David Crandall and Dima Damen and Giovanni Maria Farinella and Christian Fuegen and Bernard Ghanem and Vamsi Krishna Ithapu and C. V. Jawahar and Hanbyul Joo and Kris Kitani and Haizhou Li and Richard Newcombe and Aude Oliva and Hyun Soo Park and James M. Rehg and Yoichi Sato and Jianbo Shi and Mike Zheng Shou and Antonio Torralba and Lorenzo Torresani and Mingfei Yan and Jitendra Malik},
      year={2022},
      eprint={2110.07058},
      archivePrefix={arXiv},
      primaryClass={cs.CV},
      url={https://arxiv.org/abs/2110.07058}, 
}

@misc{miech2019howto100mlearningtextvideoembedding,
      title={HowTo100M: Learning a Text-Video Embedding by Watching Hundred Million Narrated Video Clips}, 
      author={Antoine Miech and Dimitri Zhukov and Jean-Baptiste Alayrac and Makarand Tapaswi and Ivan Laptev and Josef Sivic},
      year={2019},
      eprint={1906.03327},
      archivePrefix={arXiv},
      primaryClass={cs.CV},
      url={https://arxiv.org/abs/1906.03327}, 
}

@misc{chen2026action100mlargescalevideoaction,
      title={Action100M: A Large-scale Video Action Dataset}, 
      author={Delong Chen and Tejaswi Kasarla and Yejin Bang and Mustafa Shukor and Willy Chung and Jade Yu and Allen Bolourchi and Theo Moutakanni and Pascale Fung},
      year={2026},
      eprint={2601.10592},
      archivePrefix={arXiv},
      primaryClass={cs.CV},
      url={https://arxiv.org/abs/2601.10592}, 
}

@misc{masip2026ffjepalonghorizonplanningworld,
      title={FF-JEPA: Long-Horizon Planning in World Models with Latent Planners}, 
      author={Sergi Masip and Jonathan Swinnen and Yutong Hu and Renaud Detry and Tinne Tuytelaars},
      year={2026},
      eprint={2606.09311},
      archivePrefix={arXiv},
      primaryClass={cs.AI},
      url={https://arxiv.org/abs/2606.09311}, 
}

@misc{nair2019hierarchicalforesightselfsupervisedlearning,
      title={Hierarchical Foresight: Self-Supervised Learning of Long-Horizon Tasks via Visual Subgoal Generation}, 
      author={Suraj Nair and Chelsea Finn},
      year={2019},
      eprint={1909.05829},
      archivePrefix={arXiv},
      primaryClass={cs.LG},
      url={https://arxiv.org/abs/1909.05829}, 
}

@misc{khazatsky2025droidlargescaleinthewildrobot,
      title={DROID: A Large-Scale In-The-Wild Robot Manipulation Dataset}, 
      author={Alexander Khazatsky and Karl Pertsch and Suraj Nair and Ashwin Balakrishna and Sudeep Dasari and Siddharth Karamcheti and Soroush Nasiriany and Mohan Kumar Srirama and Lawrence Yunliang Chen and Kirsty Ellis and Peter David Fagan and Joey Hejna and Masha Itkina and Marion Lepert and Yecheng Jason Ma and Patrick Tree Miller and Jimmy Wu and Suneel Belkhale and Shivin Dass and Huy Ha and Arhan Jain and Abraham Lee and Youngwoon Lee and Marius Memmel and Sungjae Park and Ilija Radosavovic and Kaiyuan Wang and Albert Zhan and Kevin Black and Cheng Chi and Kyle Beltran Hatch and Shan Lin and Jingpei Lu and Jean Mercat and Abdul Rehman and Pannag R Sanketi and Archit Sharma and Cody Simpson and Quan Vuong and Homer Rich Walke and Blake Wulfe and Ted Xiao and Jonathan Heewon Yang and Arefeh Yavary and Tony Z. Zhao and Christopher Agia and Rohan Baijal and Mateo Guaman Castro and Daphne Chen and Qiuyu Chen and Trinity Chung and Jaimyn Drake and Ethan Paul Foster and Jensen Gao and Vitor Guizilini and David Antonio Herrera and Minho Heo and Kyle Hsu and Jiaheng Hu and Muhammad Zubair Irshad and Donovon Jackson and Charlotte Le and Yunshuang Li and Kevin Lin and Roy Lin and Zehan Ma and Abhiram Maddukuri and Suvir Mirchandani and Daniel Morton and Tony Nguyen and Abigail O'Neill and Rosario Scalise and Derick Seale and Victor Son and Stephen Tian and Emi Tran and Andrew E. Wang and Yilin Wu and Annie Xie and Jingyun Yang and Patrick Yin and Yunchu Zhang and Osbert Bastani and Glen Berseth and Jeannette Bohg and Ken Goldberg and Abhinav Gupta and Abhishek Gupta and Dinesh Jayaraman and Joseph J Lim and Jitendra Malik and Roberto Martín-Martín and Subramanian Ramamoorthy and Dorsa Sadigh and Shuran Song and Jiajun Wu and Michael C. Yip and Yuke Zhu and Thomas Kollar and Sergey Levine and Chelsea Finn},
      year={2025},
      eprint={2403.12945},
      archivePrefix={arXiv},
      primaryClass={cs.RO},
      url={https://arxiv.org/abs/2403.12945}, 
}

@misc{lecun2022path,
  title={A Path Towards Autonomous Machine Intelligence},
  author={LeCun, Yann},
  year={2022},
  note={Version 0.9.2, June 27, 2022}
}

@misc{zhang2026hierarchicalplanninglatentworld,
      title={Hierarchical Planning with Latent World Models}, 
      author={Wancong Zhang and Basile Terver and Artem Zholus and Soham Chitnis and Harsh Sutaria and Mido Assran and Randall Balestriero and Amir Bar and Adrien Bardes and Yann LeCun and Nicolas Ballas},
      year={2026},
      eprint={2604.03208},
      archivePrefix={arXiv},
      primaryClass={cs.LG},
      url={https://arxiv.org/abs/2604.03208}, 
}

@misc{hou2026worldmodelrobotlearning,
      title={World Model for Robot Learning: A Comprehensive Survey}, 
      author={Bohan Hou and Gen Li and Jindou Jia and Tuo An and Xinying Guo and Sicong Leng and Haoran Geng and Yanjie Ze and Tatsuya Harada and Philip Torr and Oier Mees and Marc Pollefeys and Zhuang Liu and Jiajun Wu and Pieter Abbeel and Jitendra Malik and Yilun Du and Jianfei Yang},
      year={2026},
      eprint={2605.00080},
      archivePrefix={arXiv},
      primaryClass={cs.RO},
      url={https://arxiv.org/abs/2605.00080}, 
}

@article{liu2023libero,
  title={LIBERO: Benchmarking Knowledge Transfer for Lifelong Robot Learning},
  author={Liu, Bo and Zhu, Yifeng and Gao, Chongkai and Feng, Yihao and Liu, Qiang and Zhu, Yuke and Stone, Peter},
  journal={arXiv preprint arXiv:2306.03310},
  year={2023}
}

@inproceedings{walke2023bridgedata,
    title={BridgeData V2: A Dataset for Robot Learning at Scale},
    author={Walke, Homer and Black, Kevin and Lee, Abraham and Kim, Moo Jin and Du, Max and Zheng, Chongyi and Zhao, Tony and Hansen-Estruch, Philippe and Vuong, Quan and He, Andre and Myers, Vivek and Fang, Kuan and Finn, Chelsea and Levine, Sergey},
    booktitle={Conference on Robot Learning (CoRL)},
    year={2023}
}

@misc{agibotworldcontributors2025agibotworldcolosseolargescale,
      title={AgiBot World Colosseo: A Large-scale Manipulation Platform for Scalable and Intelligent Embodied Systems}, 
      author={AgiBot-World-Contributors and Qingwen Bu and Jisong Cai and Li Chen and Xiuqi Cui and Yan Ding and Siyuan Feng and Shenyuan Gao and Xindong He and Xuan Hu and Xu Huang and Shu Jiang and Yuxin Jiang and Cheng Jing and Hongyang Li and Jialu Li and Chiming Liu and Yi Liu and Yuxiang Lu and Jianlan Luo and Ping Luo and Yao Mu and Yuehan Niu and Yixuan Pan and Jiangmiao Pang and Yu Qiao and Guanghui Ren and Cheng Ruan and Jiaqi Shan and Yongjian Shen and Chengshi Shi and Mingkang Shi and Modi Shi and Chonghao Sima and Jianheng Song and Huijie Wang and Wenhao Wang and Dafeng Wei and Chengen Xie and Guo Xu and Junchi Yan and Cunbiao Yang and Lei Yang and Shukai Yang and Maoqing Yao and Jia Zeng and Chi Zhang and Qinglin Zhang and Bin Zhao and Chengyue Zhao and Jiaqi Zhao and Jianchao Zhu},
      year={2025},
      eprint={2503.06669},
      archivePrefix={arXiv},
      primaryClass={cs.RO},
      url={https://arxiv.org/abs/2503.06669}, 
}

@misc{lynch2022interactivelanguagetalkingrobots,
      title={Interactive Language: Talking to Robots in Real Time}, 
      author={Corey Lynch and Ayzaan Wahid and Jonathan Tompson and Tianli Ding and James Betker and Robert Baruch and Travis Armstrong and Pete Florence},
      year={2022},
      eprint={2210.06407},
      archivePrefix={arXiv},
      primaryClass={cs.RO},
      url={https://arxiv.org/abs/2210.06407}, 
}

@misc{qwen3.5,
    title  = {{Qwen3.5}: Towards Native Multimodal Agents},
    author = {{Qwen Team}},
    month  = {February},
    year   = {2026},
    url    = {https://qwen.ai/blog?id=qwen3.5}
}

@misc{he2026demojepajointembeddingpredictivearchitecture,
      title={Demo-JEPA: Joint-Embedding Predictive Architecture for One-shot Cross-Embodiment Imitation}, 
      author={Jingyang He and Guangrun Li and Jieyu Zhang and Chengkai Hou and Zhengping Che and Shanghang Zhang},
      year={2026},
      eprint={2605.20811},
      archivePrefix={arXiv},
      primaryClass={cs.RO},
      url={https://arxiv.org/abs/2605.20811}, 
}
\bibliographystyle{iclr2027_conference}

\newpage

\appendix

\section{Environments}
\label{app_environments}

\subsection{Environment Details}

\mypar{TwoRoom} \citep{sobal2025learningrewardfreeofflinedata} is a 2D continuous-control navigation task consisting of two rooms connected by a narrow doorway. The agent is initialized in one room and tasked with reaching a target location in the other. Although goals are typically specified as images of the target state, we instead support latent goal prediction from language by expressing targets as “place the red dot at position [x, y]”. To stress-test long-horizon planning, we group test seeds into five difficulty levels based on expert trajectory length. Prior work \citep{zhou2025dinowmworldmodelspretrained, wang2026temporalstraighteninglatentplanning, maes2026leworldmodelstableendtoendjointembedding} evaluated settings where targets lay within 25 steps of the start, often yielding near-solved or trivial instances (see Appendix~\ref{tworoom_evaluation_examples} for examples). In contrast, we set each level to a fixed start-to-target distance, eliminating the risk of such degenerate cases.

\mypar{Kingdom} is a 2D navigation environment that we introduce to stress-test model-based planning and subgoal generation under dynamic spatial layouts and irreversible environmental effects. The environment is represented as a discrete grid of cells, and water cells act as terminal hazards: if the agent enters a water cell, the episode ends immediately. This introduces irreversible failure states and strengthens the need for precise long-horizon planning. At each episode reset, the positions of all entities, including the agent (a knight), four non-player characters (villager, princess, dragon, wizard), static obstacles, and water tiles, are randomly reconfigured. This ensures that neither goal locations nor feasible paths are consistent across episodes, so successful behavior cannot be memorized and must instead be inferred from the current scene configuration at each rollout. 

Kingdom defines 30 language goals with relational objectives such as “reach the left of the dragon” whose satisfaction depends jointly on the current arrangement of entities and obstacles. As in TwoRoom, we partition evaluation into multiple difficulty levels based on expert trajectory length, enabling a controlled assessment of long-horizon planning (see Appendix~\ref{kingdom_evaluation_examples} for examples).

\begin{figure}[H]
    \centering
    
    \begin{subfigure}{0.3\textwidth}
        \centering
        \includegraphics[width=\linewidth]{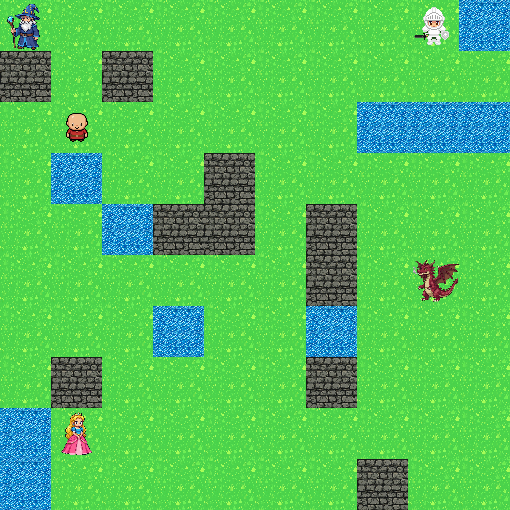} 
        \caption{Randomized initial state.}
        \label{fig:env_a}
    \end{subfigure}\hfill
    \begin{subfigure}{0.3\textwidth}
        \centering
        \includegraphics[width=\linewidth]{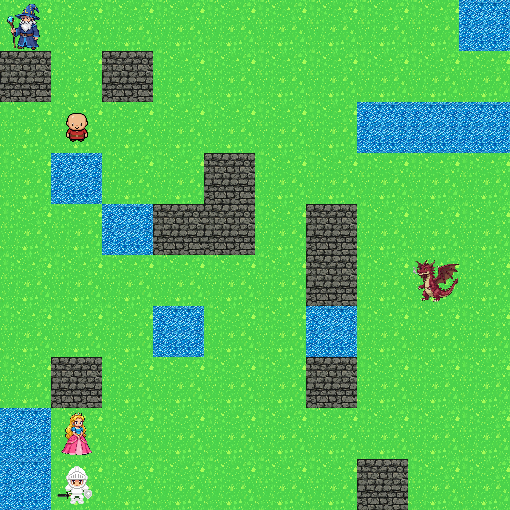} 
        \caption{``Reach below the princess''.}
        \label{fig:env_b}
    \end{subfigure}\hfill
    \begin{subfigure}{0.3\textwidth}
        \centering
        \includegraphics[width=\linewidth]{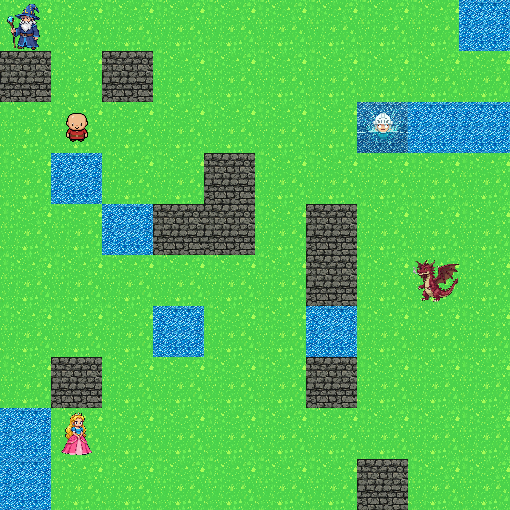} 
        \caption{Drowning failure}
        \label{fig:env_c}
    \end{subfigure}\hfill
    
    \caption{\textbf{Kingdom environment}. Example frames from randomized grid-world layouts containing walls, NPCs (villager, princess, dragon, wizard), and water. (a)~The agent at rest. (b)~A goal-satisfying state for the task ``reach below the princess'', where the agent stands on the tile directly beneath the princess. (c)~Water tiles are walkable but terminal, so a single misstep ends the episode, which makes planning difficult even at short horizons.}
    \vspace{-10pt}
    \label{fig:all_envs_grouped}
\end{figure}

\mypar{OGBench Cube} \citep{park2025ogbenchbenchmarkingofflinegoalconditioned} is a manipulation task where a robotic arm must move a cube to a target location on a tabletop. Each episode randomizes the positions of the robot, cube, and goal, preventing reliance on fixed spatial configurations. We enable latent goal prediction from language by expressing targets as “move the cube to position [x, y]”. Task difficulty is controlled by categorizing episodes according to the required displacement, defined as the sum of the end-effector-to-cube and cube-to-goal distances. Since LAGO and all planning baselines learn exclusively from pixels, grasping is difficult to predict reliably. We therefore use a simple distance-based heuristic that automatically closes the gripper when the end-effector is within a specified distance of the cube. This heuristic is applied to all model-based planning methods.

\mypar{OGBench Scene} \citep{park2025ogbenchbenchmarkingofflinegoalconditioned} is a manipulation environment with three interactable objects: a drawer (which can be opened, closed, locked, or unlocked via a button), a window (similarly button-controlled), and a cube that can be rearranged or placed inside the drawer. Unlike Kingdom or OGBench Cube, the scene layout is largely fixed so goal localization is static and the planning challenge lies primarily in task ordering and execution rather than spatial reasoning.

\subsection{Evaluation Examples}
\label{evaluation_examples}

\subsubsection{TwoRoom}
\label{tworoom_evaluation_examples}
To illustrate the difference between prior evaluation protocols and our difficulty-graded benchmark, we provide randomly selected test configurations from both settings below.

\begin{figure}[H]
    \centering
    \includegraphics[width=\textwidth]{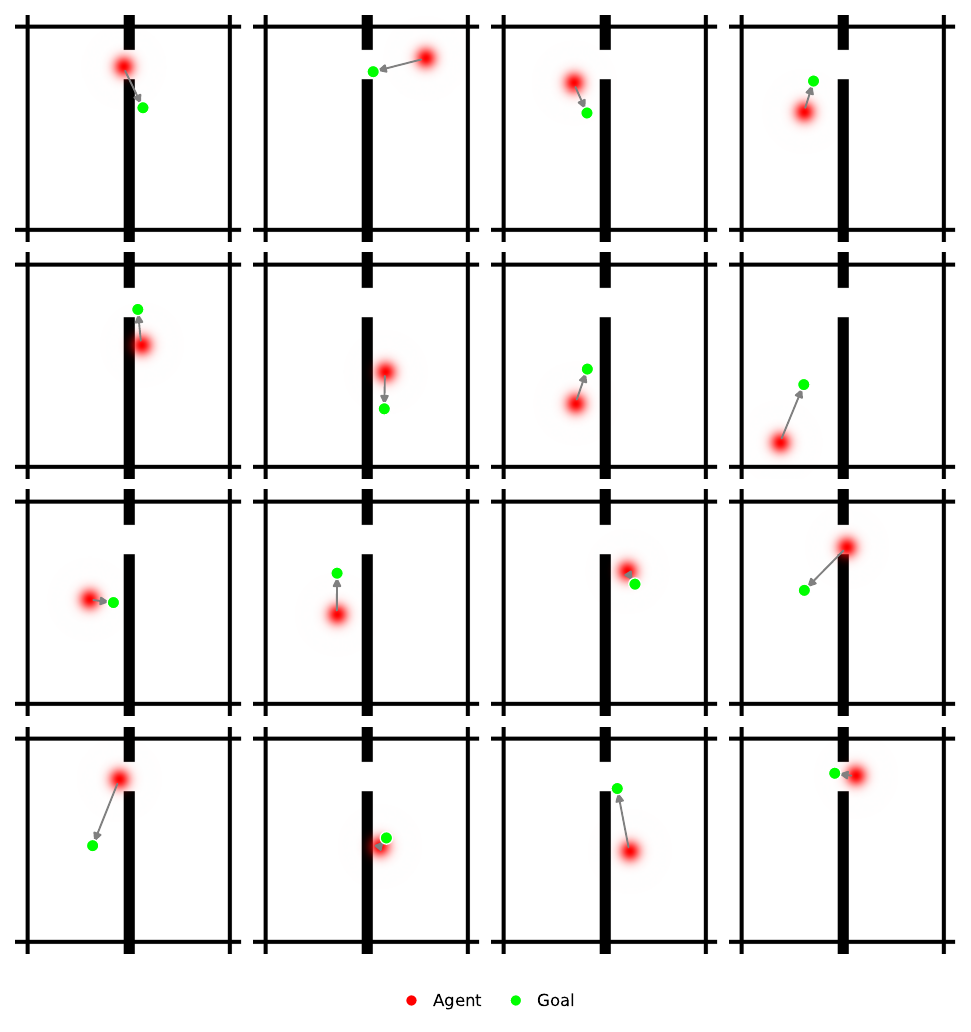} 
    \caption{Randomly selected examples of prior TwoRoom test configurations, where start (\emph{red}) and goal (\emph{green}) states are configured to be at most 25 steps apart, often resulting in near-trivial or already-solved instances.}
    \label{fig:supp_tworoom_prior_examples}
\end{figure}

\begin{figure}[H]
    \centering
    \includegraphics[width=\textwidth]{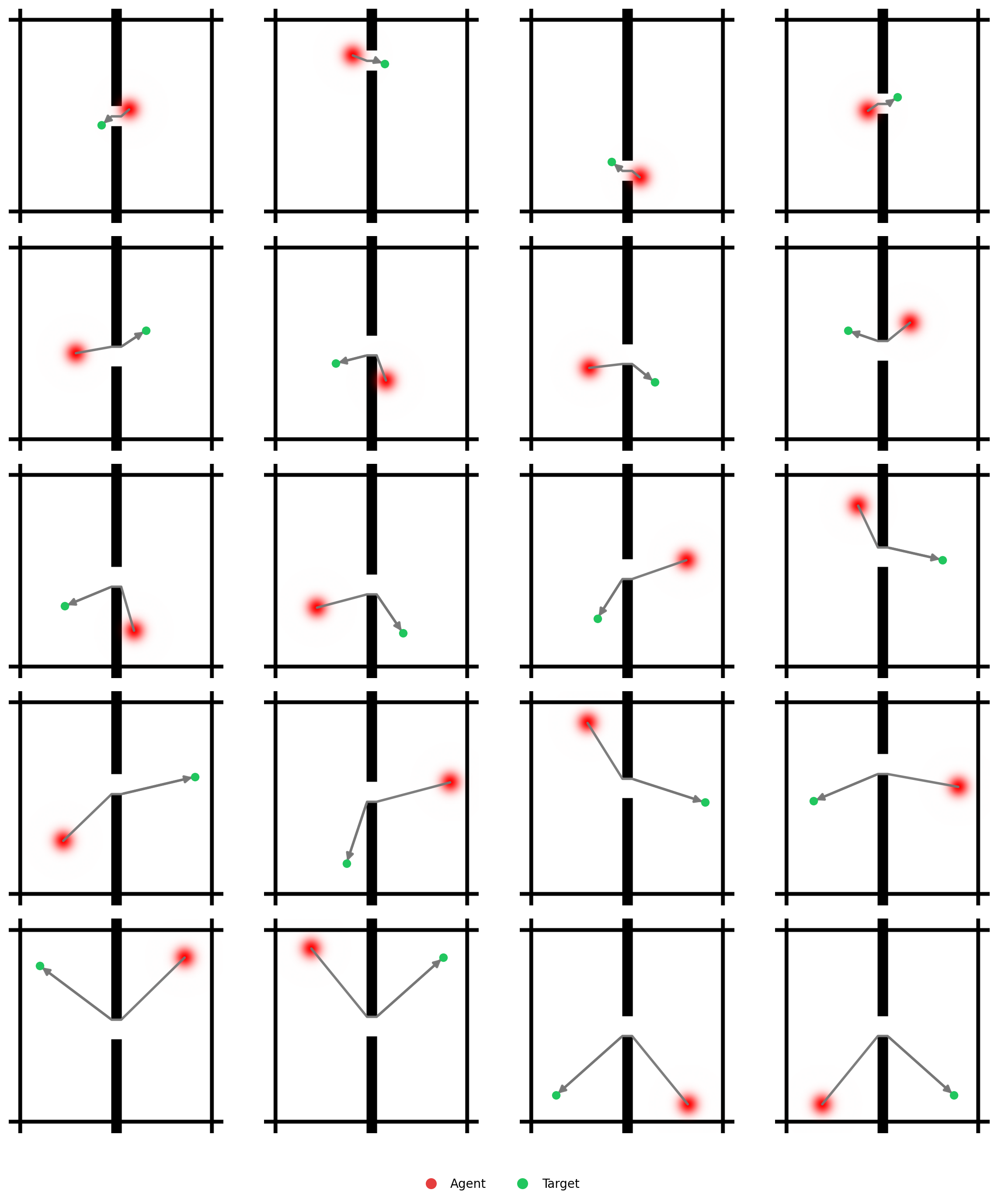} 
    \caption{Randomly selected examples of our difficulty-graded TwoRoom test configurations. Each row represents a different difficulty level. While previous test suites only enforce a maximum distance, we enforce an exact distance between the starting and target positions to better evaluate the long-horizon capacity of model-based planning.}
    \label{fig:supp_tworoom_graded_examples1}
\end{figure}

\subsubsection{Kingdom}
\label{kingdom_evaluation_examples}

\begin{figure}[H]
    \centering
    \includegraphics[width=\textwidth]{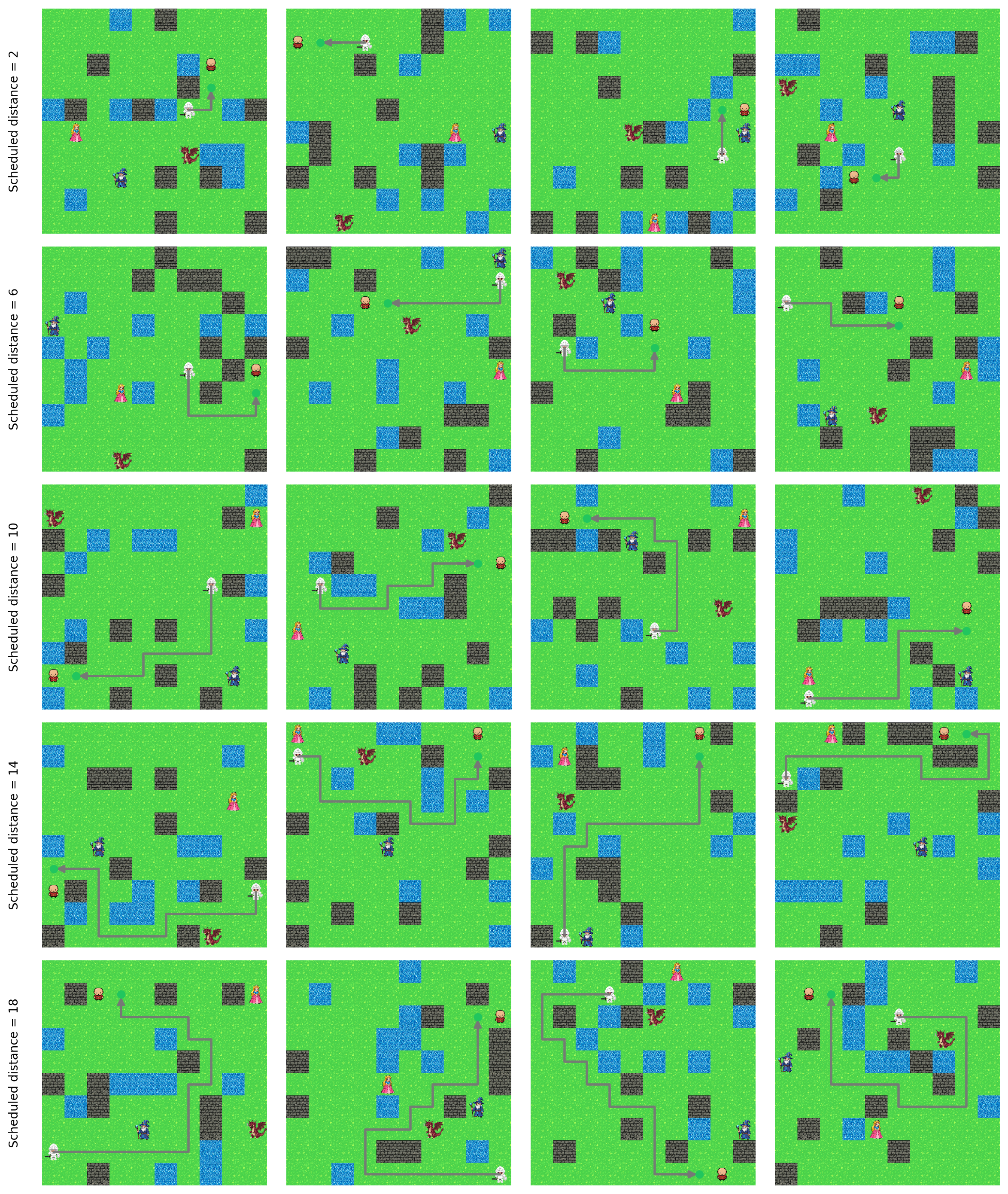} 
    \caption{Randomly selected examples of our difficulty-graded Kingdom test configurations. Each row represents a different difficulty level based on the distance to travel to reach the target. In these examples, the language goal is "Reach the villager".}
    \label{fig:supp_tworoom_graded_examples2}
\end{figure}

\section{Complete results tables}
\label{app:complete_results_table}

\subsection{Subgoal performance}

\begin{table}[H]
  \caption{Complete results for all environments for increasing levels of difficulty, averaged across all possible tasks for an environment. Values represent mean success rate $\pm$ std, where higher is better.}
  \label{tab:full-results}
  \centering
  \resizebox{\linewidth}{!}{%
  \begin{tabular}{lcccccccc}
    \toprule[\heavyrulewidth]
    Difficulty & DINO-WM & DINO-WM (O.S.) & LeWM & T.S. & HWM & LAGO (ours) & LAGO (O.S.) & GCBC \\
    \midrule
    \multicolumn{9}{c}{\textit{TwoRoom}} \\
    \midrule
    35  & 100.0\ppm{0.0} & 100.0\ppm{0.0} & 20.0\ppm{12.6} & 100.0\ppm{0.0} & 100.0\ppm{0.0} & 100.0\ppm{0.0} & 100.0\ppm{0.0} & 100.0\ppm{0.0} \\
    75  & 10.0\ppm{9.5} & 100.0\ppm{0.0} & 0.0\ppm{0.0} & 90.0\ppm{9.5} & 50.0\ppm{7.1} & 100.0\ppm{0.0} & 100.0\ppm{0.0} & 100.0\ppm{0.0} \\
    115 & 10.0\ppm{9.5} & 100.0\ppm{0.0} & 0.0\ppm{0.0} & 0.0\ppm{0.0} & 44.0\ppm{7.0} & 100.0\ppm{0.0} & 100.0\ppm{0.0} & 100.0\ppm{0.0} \\
    155 & 0.0\ppm{0.0} & 100.0\ppm{0.0} & 0.0\ppm{0.0} & 0.0\ppm{0.0} & 16.0\ppm{5.2} & 100.0\ppm{0.0} & 100.0\ppm{0.0} & 100.0\ppm{0.0} \\
    200 & 0.0\ppm{0.0} & 100.0\ppm{0.0} & 0.0\ppm{0.0} & 0.0\ppm{0.0} & 0.0\ppm{0.0} & 100.0\ppm{0.0} & 100.0\ppm{0.0} & 100.0\ppm{0.0} \\
    \midrule
    \multicolumn{9}{c}{\textit{Kingdom}} \\
    \midrule
    2  & 97.0\ppm{1.0} & 100.0\ppm{0.0} & 10.0\ppm{1.7} & 98.0\ppm{0.8} & 83.4\ppm{2.2} & 100.0\ppm{0.0} & 100.0\ppm{0.0} & 89.7\ppm{2.5} \\
    6  & 29.0\ppm{2.7} & 100.0\ppm{0.0} & 2.0\ppm{0.9} & 78.0\ppm{2.4} & 10.7\ppm{1.8} & 97.0\ppm{1.0} & 100.0\ppm{0.0} & 55.9\ppm{4.1} \\
    10 & 10.0\ppm{0.5} & 88.0\ppm{1.9} & 0.0\ppm{0.0} & 15.0\ppm{2.1} & 1.0\ppm{0.6} & 84.0\ppm{2.2} & 98.0\ppm{0.9} & 29.7\ppm{3.8} \\
    14 & 0.0\ppm{0.0} & 46.0\ppm{2.9} & 0.0\ppm{0.0} & 0.0\ppm{0.0} & 0.3\ppm{0.3} & 39.0\ppm{2.9} & 76.0\ppm{2.5} & 9.7\ppm{2.5} \\
    18 & 0.0\ppm{0.0} & 15.0\ppm{2.1} & 0.0\ppm{0.0} & 0.0\ppm{0.0} & 0.0\ppm{0.0} & 8.0\ppm{1.6} & 39.0\ppm{2.9} & 1.4\ppm{1.0} \\
    \midrule
    \multicolumn{9}{c}{\textit{OGBench Cube}} \\
    \midrule
    0.08 & 80.0\ppm{4.0} & 99.0\ppm{1.0} & 34.0\ppm{4.7} & 75.0\ppm{4.3} & 86.0\ppm{3.5} & 85.0\ppm{3.6} & 94.0\ppm{3.4} & 70.0\ppm{12.6} \\
    0.12 & 24.0\ppm{4.3} & 57.0\ppm{5.0} & 6.0\ppm{2.4} & 32.0\ppm{4.7} & 43.0\ppm{5.0} & 60.0\ppm{4.9} & 74.0\ppm{6.2} & 35.0\ppm{10.7} \\
    0.2  & 3.0\ppm{1.7} & 44.0\ppm{5.0} & 0.0\ppm{0.0} & 9.0\ppm{2.9} & 7.0\ppm{2.6} & 55.0\ppm{5.0} & 46.0\ppm{7.0} & 45.0\ppm{11.1} \\
    0.3  & 0.0\ppm{0.0} & 20.0\ppm{4.0} & 0.0\ppm{0.0} & 0.0\ppm{0.0} & 6.0\ppm{2.4} & 29.0\ppm{4.5} & 32.0\ppm{6.6} & 45.0\ppm{11.1} \\
    0.4  & 0.0\ppm{0.0} & 9.0\ppm{2.9} & 0.0\ppm{0.0} & 0.0\ppm{0.0} & 0.0\ppm{0.0} & 11.0\ppm{3.1} & 20.0\ppm{6.0} & 40.0\ppm{10.9} \\
    \bottomrule
  \end{tabular}%
  }
\end{table}

\subsection{Language grounding}

\begin{table}[H]
  \caption{Language goal performance across environments. Values are mean success rate $\pm$ std, where higher is better.}
  \label{tab:language-performance}
  \centering
  \small
  \begin{tabular}{lc}
    \toprule[\heavyrulewidth]
    Method & Success Rate (\%) \\
    \midrule
    \multicolumn{2}{c}{\textit{Kingdom}} \\
    \midrule
    PE-Core & 6.00\ppm{0.93} \\
    LIV & 8.77\ppm{1.11} \\
    LIV (finetuned) & 52.6\ppm{1.38} \\
    LAGO (ours) & 97.0\ppm{1.0} \\
    \midrule
    \multicolumn{2}{c}{\textit{OGBench Scene}} \\
    \midrule
    PE-Core & 0.0\ppm{0.0} \\
    LIV & 12.5\ppm{5.2} \\
    LIV (finetuned) & 48.75\ppm{5.6} \\
    LAGO (ours) & 55.31\ppm{2.36} \\
    \bottomrule
  \end{tabular}
\end{table}

\subsection{Performance per Number of Subgoals}

\begin{table}[H]
  \caption{Complete results of LAGO's subgoal-count ablation for each environment. TwoRoom is fixed at difficulty 155, and Kingdom at difficulty 10. Values are success rates in $[0, 1]$, where higher is better.}
  \label{tab:num-subgoals-full-results}
  \centering
  \setlength{\tabcolsep}{4.5pt}
  \resizebox{\columnwidth}{!}{
  \begin{tabular}{lccccccccc}
    \toprule[\heavyrulewidth]
    & \multicolumn{9}{c}{Number of subgoals} \\
    \cmidrule(lr){2-10}
    Environment & 1 & 2 & 3 & 4 & 5 & 6 & 7 & 8 & 10 \\
    \midrule
    TwoRoom & 0.00\ppm{0.00} & 0.92\ppm{0.05} & 0.98\ppm{0.02} & \textbf{1.00\ppm{0.00}} & \textbf{1.00\ppm{0.00}} & \textbf{1.00\ppm{0.00}} & \textbf{1.00\ppm{0.00}} & \textbf{1.00\ppm{0.00}} & \textbf{1.00\ppm{0.00}} \\
    Kingdom & 0.00\ppm{0.00} & 0.18\ppm{0.09} & 0.40\ppm{0.15} & 0.70\ppm{0.10} & 0.78\ppm{0.13} & 0.74\ppm{0.14} & 0.80\ppm{0.11} & 0.78\ppm{0.12} & \textbf{0.84\ppm{0.11}} \\
    \midrule
    Average & 0.00 & 0.55 & 0.69 & 0.85 & 0.89 & 0.87 & 0.90 & 0.89 & \textbf{0.92} \\
    \bottomrule
  \end{tabular}%
  }
\end{table}

\subsection{Performance per Cost Temperature}
\label{app:cost}

\begin{table}[H]
  \caption{Complete results of LAGO's cost function analysis for each environment. TwoRoom is fixed at difficulty 155, and Kingdom at difficulty 10. Soft-min is reported for temperatures $\tau \in \{0.001, 0.01, 0.03, 0.1, 0.3, 1\}$. Values are success rates in \%, $\pm$ std, where higher is better.}
  \label{tab:cost-function-full-results}
  \centering
  \resizebox{\linewidth}{!}{%
  \begin{tabular}{lcccccccc}
    \toprule[\heavyrulewidth]
    & \multirow{2}{*}{Sum} & \multicolumn{6}{c}{Soft-Min ($\tau$)} \\
    \cmidrule(lr){3-8}
    Environment & & 0.001 & 0.01 & 0.03 & 0.1 & 0.3 & 1 \\
    \midrule
    TwoRoom & 50.0\ppm{16.0} & \textbf{100.0}\ppm{0.0} & \textbf{100.0}\ppm{0.0} & \textbf{100.0}\ppm{0.0} & \textbf{100.0}\ppm{0.0} & \textbf{100.0}\ppm{0.0} & 80.0\ppm{13.0} \\
    Kingdom & 0.0\ppm{0.0}   & 70.0\ppm{14.0}          & \textbf{80.0}\ppm{13.0} & 60.0\ppm{15.0}          & 10.0\ppm{9.0}           & 0.0\ppm{0.0}            & 0.0\ppm{0.0} \\
    \midrule
    Average & 25.0\ppm{8.0}  & 85.0\ppm{7.0}           & \textbf{90.0}\ppm{6.5}  & 80.0\ppm{7.5}           & 55.0\ppm{4.5}           & 50.0\ppm{0.0}           & 40.0\ppm{6.5} \\
    \bottomrule
  \end{tabular}%
  }
\end{table}

\newpage

\section{Hyperparameters}
\label{hyperparameters}

\subsection{LAGO}

\begin{table}[H]
  \centering
  \caption{Architectural hyperparameters of LAGO.}
  \label{tab:arch-hyperparameters}
  \small
  \begin{tabular}{lcc}
    \toprule
    \textbf{Parameter} & \textbf{Symbol} & \textbf{Value} \\
    \midrule
    Total parameters          &         & 15.0M \\
    \midrule
    \multicolumn{3}{l}{\textit{Vision \& Language Encoders}} \\[2pt]
    Patch token dimension     & $D_z$   & 384 \\
    Language embedding dim.   & $D_L$   & 768 \\
    Vision encoder backbone          &         & DINOv2 ViT-S/14\\
    Language encoder backbone          &         & EmbeddingGemma-300M\\
    \midrule
    \multicolumn{3}{l}{\textit{ViT Predictor}} \\[2pt]
    Depth                     &         & 6 \\
    Number of heads           &         & 8 \\
    Conditioning dimension    & $D_c$   & 64 \\
    Hidden dimension          &         & 448 \\
    Activation                &         & GELU \\
    \midrule
    \multicolumn{3}{l}{\textit{Optimisation}} \\[2pt]
    Optimizer                 &         & AdamW \\
    Learning rate             &         & $3 \times 10^{-4}$ \\
    Batch size                &         & 128 \\
    Training steps            &         & 500 epochs \\
    Early stopping patience   &         & 50 epochs \\
    \bottomrule
  \end{tabular}
\end{table}

\subsection{DINO-WM}

\begin{table}[H]
  \centering
  \caption{Architectural hyperparameters of DINO-WM.}
  \label{tab:dinowm-arch-hyperparameters}
  \small
  \begin{tabular}{lcc}
    \toprule
    \textbf{Parameter} & \textbf{Symbol} & \textbf{Value} \\
    \midrule
    Total parameters          &         & 14.7M \\
    \midrule
    \multicolumn{3}{l}{\textit{Vision Encoder}} \\[2pt]
    Patch token dimension     & $D_z$   & 384 \\
    Vision encoder backbone          &         & DINOv2 ViT-S/14\\
    \midrule
    \multicolumn{3}{l}{\textit{ViT Predictor}} \\[2pt]
    Depth                     &         & 6 \\
    Number of heads           &         & 8 \\
    Conditioning dimension    & $D_c$   & 64 \\
    Hidden dimension          &         & 448 \\
    Activation                &         & GELU \\
    \midrule
    \multicolumn{3}{l}{\textit{Optimisation}} \\[2pt]
    Optimizer                 &         & AdamW \\
    Learning rate             &         & $3 \times 10^{-4}$ \\
    Batch size                &         & 128 \\
    Training steps            &         & 500 epochs \\
    Early stopping patience   &         & 50 epochs \\
    \bottomrule
  \end{tabular}
\end{table}

\subsection{Temporal Straightening}

\begin{table}[H]
  \centering
  \caption{Architectural hyperparameters of Temporal Straightening.}
  \label{tab:arch-hyperparameters-tempstraight}
  \small
  \begin{tabular}{lcc}
    \toprule
    \textbf{Parameter} & \textbf{Symbol} & \textbf{Value} \\
    \midrule
    Total parameters          &         & 17.9M \\
    \midrule
    \multicolumn{3}{l}{\textit{Vision Encoder}} \\[2pt]
    Patch token dimension     & $D_z$   & 384 \\
    Encoder backbone          &         & DINOv2 ViT-S/14\\
    \midrule
    \multicolumn{3}{l}{\textit{ViT Predictor}} \\[2pt]
    Depth                     &         & 6 \\
    Number of heads           &         & 8 \\
    Conditioning dimension    & $D_c$   & 64 \\
    Hidden dimension          &         & 448 \\
    Activation                &         & GELU \\
    \midrule
    \multicolumn{3}{l}{\textit{Straightening Aggregator}} \\[2pt]
    Output dimension          &         & 128 \\
    Hidden dimension          &         & 256 \\
    Depth                     &         & 2 \\
    \midrule
    \multicolumn{3}{l}{\textit{Straightening Loss}} \\[2pt]
    Loss weight               & $\lambda_{\mathrm{str}}$ & 0.1 \\
    \midrule
    \multicolumn{3}{l}{\textit{Optimisation}} \\[2pt]
    Optimizer                 &         & AdamW \\
    Predictor learning rate   &         & $3 \times 10^{-4}$ \\
    Projector/aggregator learning rate &  & $1 \times 10^{-5}$ \\
    Batch size                &         & 128 \\
    Training steps            &         & 500 epochs \\
    Early stopping patience   &         & 50 epochs \\
    \bottomrule
  \end{tabular}
\end{table}

\subsection{LeWM}

\begin{table}[H]
  \centering
  \caption{Architectural hyperparameters of LeWM.}
  \label{tab:arch-hyperparameters-lewm}
  \small
  \begin{tabular}{lcc}
    \toprule
    \textbf{Parameter} & \textbf{Symbol} & \textbf{Value} \\
    \midrule
    Total parameters          &         & 11.9M \\
    \midrule
    \multicolumn{3}{l}{\textit{Vision Encoder}} \\[2pt]
    Patch token dimension     & $D_z$   & 192 \\
    Encoder backbone          &         & ViT-Tiny \textsuperscript{*} \\
    \midrule
    \multicolumn{3}{l}{\textit{ViT Predictor}} \\[2pt]
    Depth                     &         & 6 \\
    Number of heads           &         & 8 \\
    Hidden dimension          &         & 256 \\
    Conditioning dimension    & $D_c$   & 64 \\
    Activation                &         & GELU \\
    \midrule
    \multicolumn{3}{l}{\textit{SIGReg}} \\[2pt]
    Loss weight               &         & 0.09 \\
    Number of knots           &         & 17 \\
    Number of projections     &         & 1024 \\
    \midrule
    \multicolumn{3}{l}{\textit{Optimisation}} \\[2pt]
    Optimizer                 &         & AdamW \\
    Learning rate             &         & $5 \times 10^{-5}$ \textsuperscript{*} \\
    Batch size                &         & 128 \\
    Training steps            &         & 500 epochs \\
    Early stopping patience   &         & 50 epochs \\
    \bottomrule
  \end{tabular}
  \\[4pt]
  {\small \textsuperscript{*}\ We choose these values for LeWM based on the official manuscript and implementation.}\\
\end{table}

\subsection{GCBC}

\begin{table}[H]
  \centering
  \caption{Architectural hyperparameters of GCBC.}
  \label{tab:arch-hyperparameters-bc}
  \small
  \begin{tabular}{lcc}
    \toprule
    \textbf{Parameter} & \textbf{Symbol} & \textbf{Value} \\
    \midrule
    Total parameters          &         & 15.0M \\
    \midrule
    \multicolumn{3}{l}{\textit{Vision \& Language Encoders}} \\[2pt]
    Patch token dimension     & $D_z$   & 384 \\
    Language embedding dim.   & $D_L$   & 768 \\
    Vision encoder backbone          &         & DINOv2 ViT-S/14\\
    Language encoder backbone          &         & EmbeddingGemma-300M\\
    \midrule
    \multicolumn{3}{l}{\textit{ViT Predictor}} \\[2pt]
    Depth                     &         & 6 \\
    Number of heads           &         & 8 \\
    Hidden dimension          &         & 448 \\
    Conditioning dimension    & $D_c$   & 64 \\
    Activation                &         & GELU \\
    \midrule
    \multicolumn{3}{l}{\textit{Optimisation}} \\[2pt]
    Optimizer                 &         & AdamW \\
    Learning rate             &         & $3 \times 10^{-4}$ \\
    Batch size                &         & 128 \\
    Training steps            &         & 500 epochs \\
    Early stopping patience   &         & 50 epochs \\
    Frameskip                 &         & 1 \\
    \bottomrule
  \end{tabular}
\end{table}

\subsection{HWM}

\begin{table}[H]
  \centering
  \caption{Architectural hyperparameters of HWM.}
  \label{tab:arch-hyperparameters-hwm}
  \small
  \begin{tabular}{lcc}
    \toprule
    \textbf{Parameter} & \textbf{Symbol} & \textbf{Value} \\
    \midrule
    Total parameters          &         & 35.7M (sum of both levels) \\
    \midrule
    \multicolumn{3}{l}{\textit{Vision Encoder (shared, frozen)}} \\[2pt]
    Patch token dimension     & $D_z$   & 384 \\
    Encoder backbone          &         & DINOv2 ViT-S/14 \\
    \midrule
    \multicolumn{3}{l}{\textit{Level-1 Predictor (Conv)}} \\[2pt]
    Depth                     &         & 3 \\
    Channels                  &         & 512 \\
    Kernel size               &         & 3 \\
    Activation                &         & ReLU \\
    Action conditioning       &         & raw action (concat) \\
    \midrule
    \multicolumn{3}{l}{\textit{Level-2 Predictor (Conv)}} \\[2pt]
    Depth                     &         & 5 \\
    Channels                  &         & 512 \\
    Kernel size               &         & 5 \\
    Activation                &         & ReLU \\
    Macro-action latent dim.  & $D_z^{(2)}$ & 8 \\
    Action encoder            &         & 8--64--8 MLP \\
    \midrule
    \multicolumn{3}{l}{\textit{Level-2 Posterior}} \\[2pt]
    Input dimension           &         & 25 \\
    Hidden dimension          &         & 32 \\
    Depth                     &         & 2 \\
    \midrule
    \multicolumn{3}{l}{\textit{Hierarchy}} \\[2pt]
    Level-1 window length     &         & 15 steps \\
    Level-2 macro-step skip   &         & 5 \\
    Level-2 macro-steps       &         & 5 \\
    \midrule
    \multicolumn{3}{l}{\textit{Optimisation}} \\[2pt]
    Optimizer                 &         & Adam \\
    Learning rate             &         & $1.5 \times 10^{-4}$ \textsuperscript{*} \\
    Batch size                &         & 128 \\
    Training steps            &         & 500 epochs \\
    Early stopping patience   &         & 50 epochs \\
    Frameskip                 &         & 1 \\
    \bottomrule
  \end{tabular}
  \\[4pt]
  {\small \textsuperscript{*}\ The official implementation scales the configured base rate of $3 \times 10^{-4}$ by $\text{batch size}/256$, after a linear warmup over the first 10\% of steps.}\\
\end{table}

\subsection{MPC and CEM}

\begin{table}[H]
  \centering
  \caption{MPC and CEM hyperparameters used at planning time.}
  \label{tab:planning-hyperparameters}
  \begin{tabular}{lcc}
    \toprule
    \textbf{Parameter} & \textbf{Symbol} & \textbf{Value} \\
    \midrule
    Planning horizon       &    &  10\\
    Number of samples      &     &  300\\
    Number of elites       &    &  30\\
    CEM iterations            &           &  10\\
    Number of subgoals        & $K$       &  5\\
    Subgoal decay factor      & $\lambda$ &  1\\
    Subgoal decay factor (OGBench Cube)     & $\lambda$ &  0.75\\
    Temperature               & $\tau$    &  0.001\\
    \bottomrule
  \end{tabular}
\end{table}

\section{Visualizing the Loss Landscapes}
\label{app:vis_loss_landscape}

\begin{figure}[H]
    \centering
    
    \begin{subfigure}{0.4\textwidth}
        \centering
        \includegraphics[width=\linewidth]{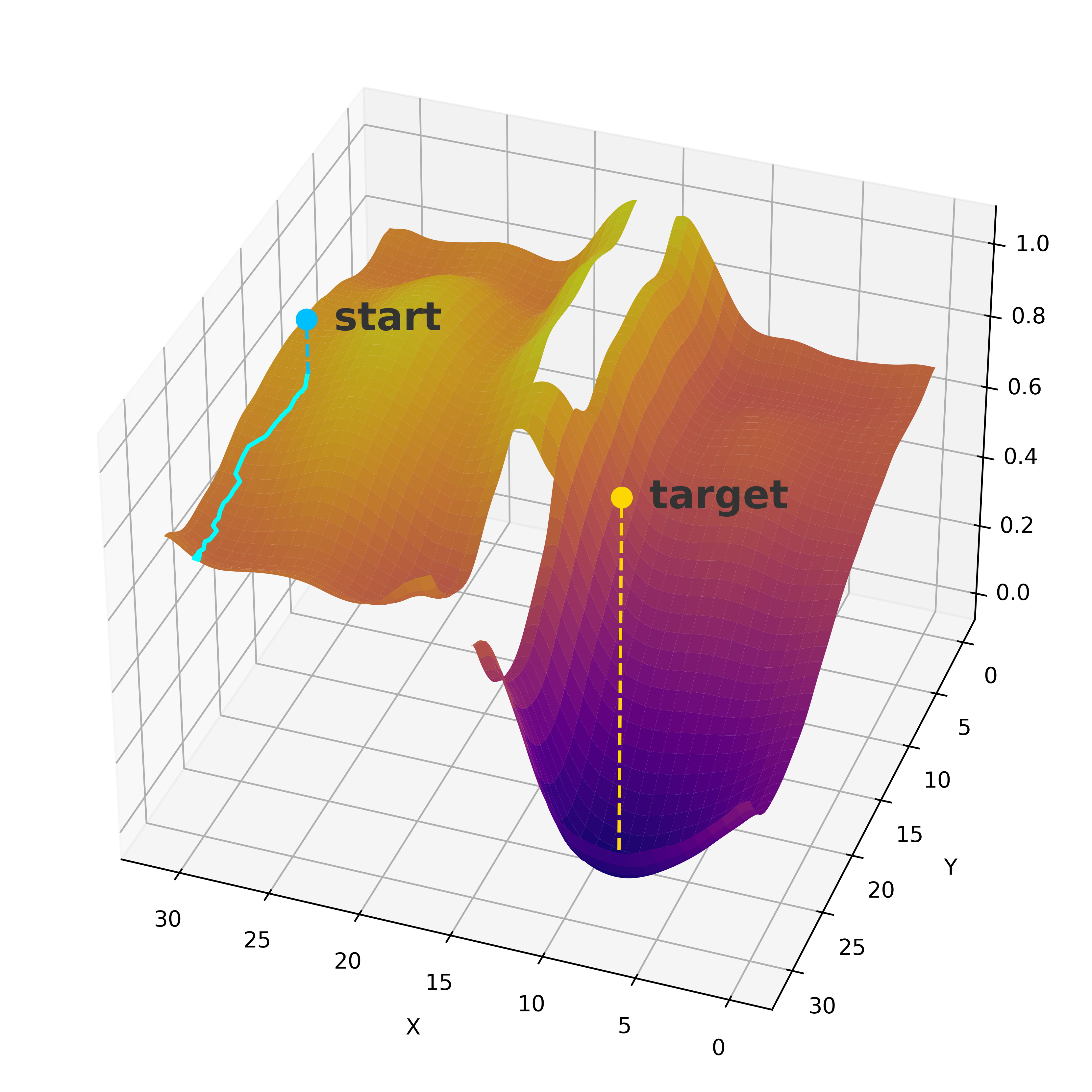} 
        \caption{Final goal image}
        \label{fig:env_a_tworoom}
    \end{subfigure}\hspace{20pt}
    \begin{subfigure}{0.4\textwidth}
        \centering
        \includegraphics[width=\linewidth]{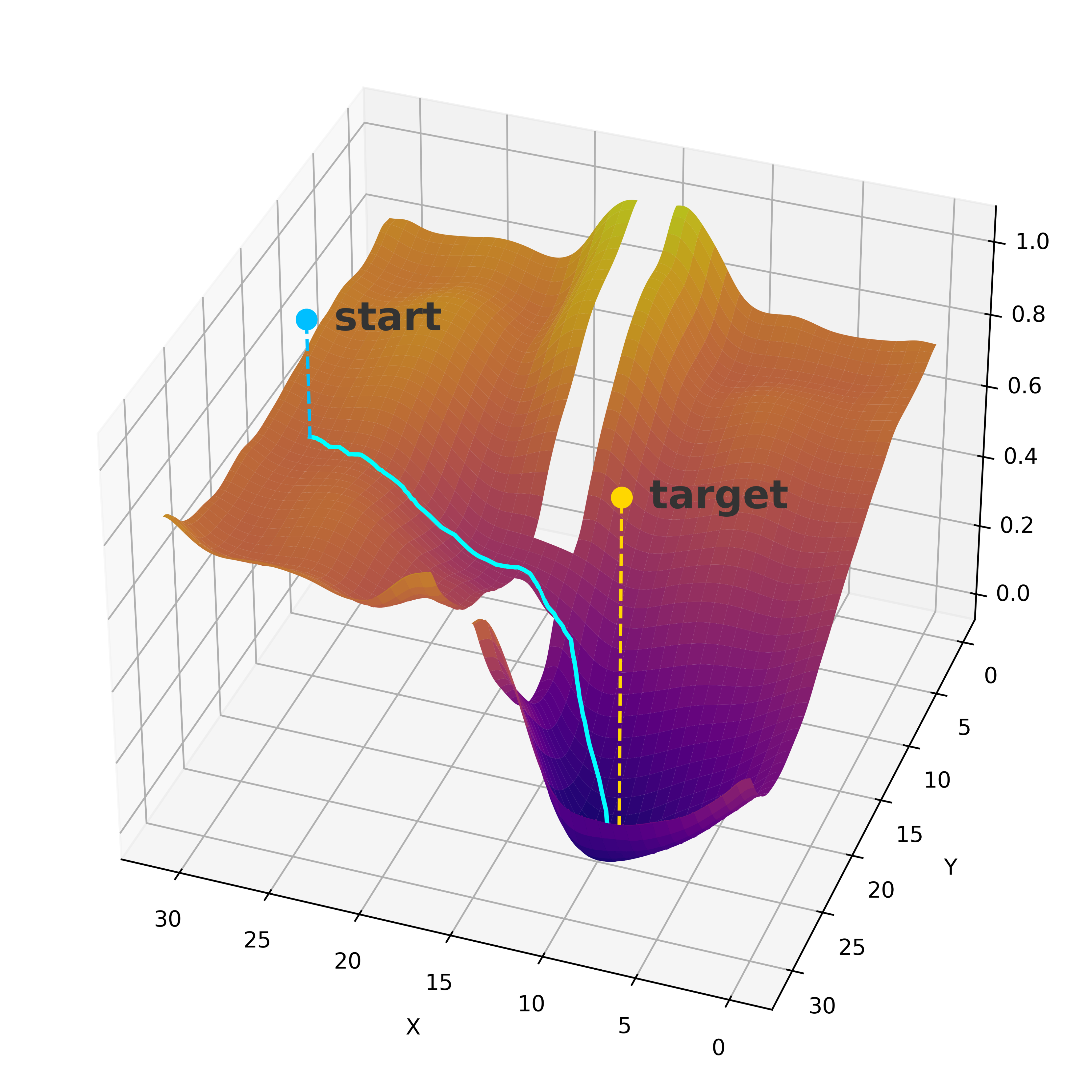} 
        \caption{LAGO}
        \label{fig:env_c_tworoom}
    \end{subfigure}
    \caption{Loss landscapes for long-horizon TwoRoom tasks: single latent goal (left) vs.\ $K$ language-predicted subgoals aggregated via $J_{\text{LAGO}}$ (right). Planned paths are shown in blue. The single-goal landscape provides little gradient for the planner, yielding suboptimal trajectories. LAGO's subgoals carve a smooth valley that guides the CEM optimizer efficiently toward the objective.}
    \label{fig:loss_landscapes}
\end{figure}

Figure~\ref{fig:loss_landscapes} visualizes the optimization landscape induced by the planning objective on a long-horizon TwoRoom task. With only the final goal as supervision, the landscape provides little useful gradient toward the goal, making optimization difficult. In contrast, LAGO's intermediate subgoals create a smooth sequence of lower-cost regions that forms a navigable valley toward the objective, providing CEM with progressively informative guidance throughout the rollout.

\section{Examples of decoded subgoals}
\label{decoded_subgoals}

We provide qualitative examples of decoded LAGO subgoals across all four environments. Each row shows the starting state (leftmost), the decoded subgoals at increasing completion values (middle), and the target state (rightmost), illustrating how LAGO decomposes a language goal into a smooth sequence of intermediate states.

\begin{figure}[H]
    \centering
    \begin{subfigure}{\textwidth}
        \centering
        \includegraphics[width=\textwidth]{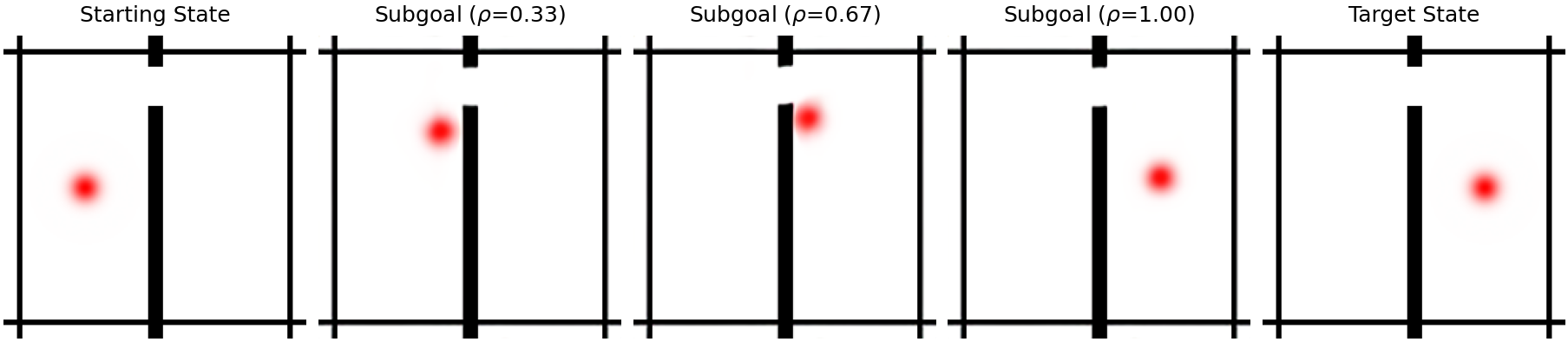}
        \caption{Example of predicted subgoals for the TwoRoom environment where the language goal passed to LAGO is "Place the red dot at position [0.732, 0.500]". The leftmost image shows the starting state, the rightmost image the target state, and middle images the decoded subgoals at different completion values.}
        \label{fig:supp_tworoom_examples}
    \end{subfigure}
    \vspace{1em}
    \begin{subfigure}{\textwidth}
        \centering
        \includegraphics[width=\textwidth]{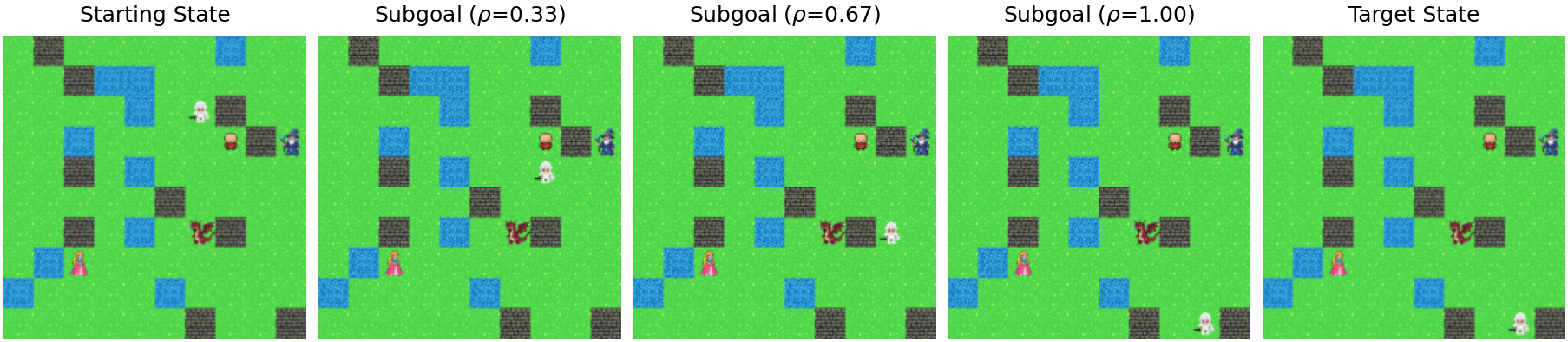}
        \caption{Example of predicted subgoals for the Kingdom environment where the language goal passed to LAGO is "Reach the bottom-right corner". The leftmost image shows the starting state, the rightmost image the target state, and middle images the decoded subgoals at different completion values.}
        \label{fig:supp_kingdom_examples}
    \end{subfigure}
    \vspace{1em}
    \begin{subfigure}{\textwidth}
        \centering
        \includegraphics[width=\textwidth]{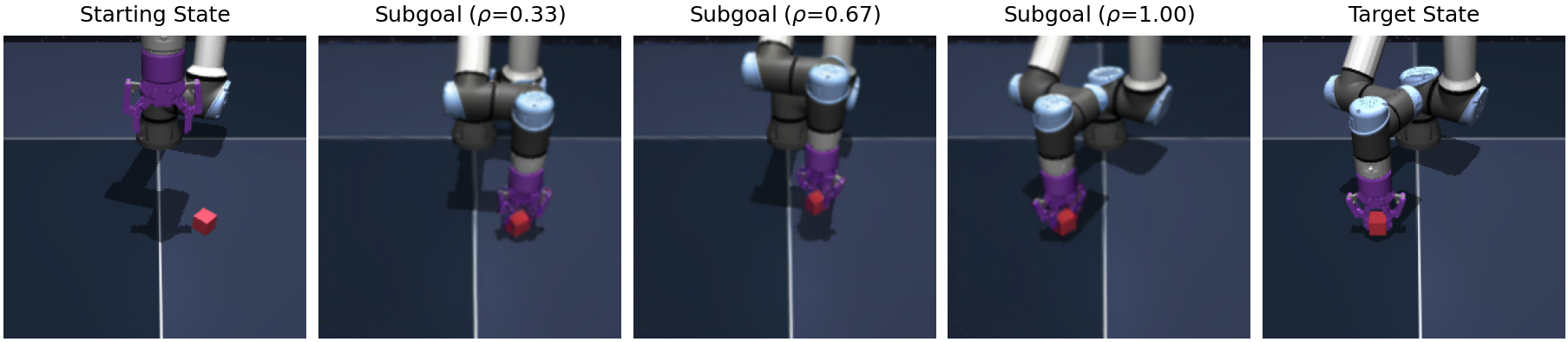}
        \caption{Example of predicted subgoals for the OGBench Cube environment where the language goal passed to LAGO is "Move the cube to position [0.425, -0.100]". The leftmost image shows the starting state, the rightmost image the target state, and middle images the decoded subgoals at different completion values.}
        \label{fig:supp_ogbcube_examples}
    \end{subfigure}
    \vspace{1em}
    \begin{subfigure}{\textwidth}
        \centering
        \includegraphics[width=\textwidth]{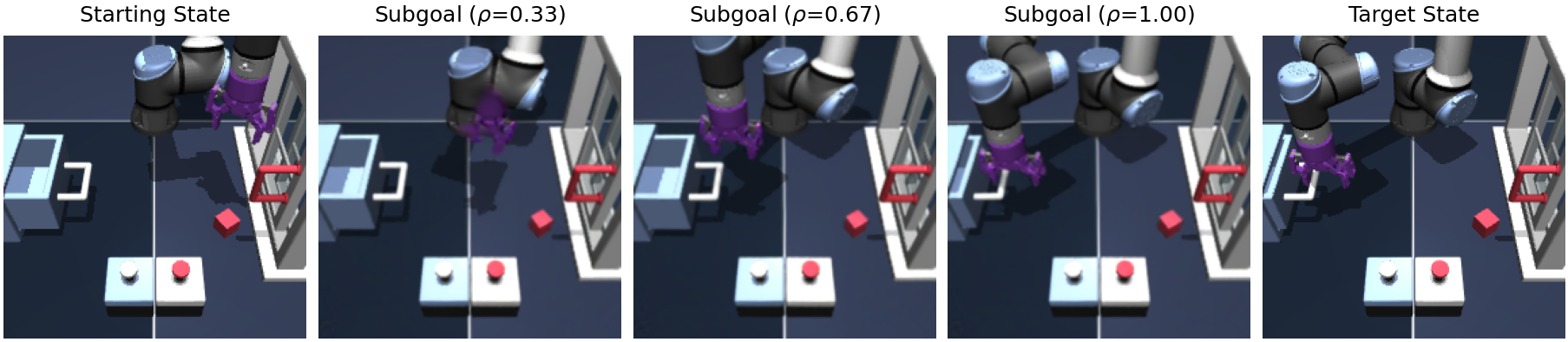}
        \caption{Example of predicted subgoals for the OGBench Scene environment where the language goal passed to LAGO is "Close the drawer". The leftmost image shows the starting state, the rightmost image the target state, and middle images the decoded subgoals at different completion values.}
        \label{fig:supp_ogbscene_examples}
    \end{subfigure}
\end{figure}

\section{Latent Space Visualization}
\label{latent_space_visualization}

To evaluate whether the learned latent spaces of the considered baselines capture sufficient spatial and semantic information, we train a lightweight decoder network to reconstruct the original pixels from the frozen latent embeddings. Since JEPA-based architectures do not inherently possess a generative component, this probing mechanism allows us to qualitatively assess the fidelity of the world model’s internal states.

\subsection{LeWM}

\mypar{TwoRoom}

\begin{figure}[H]
    \centering
    \includegraphics[width=\textwidth]{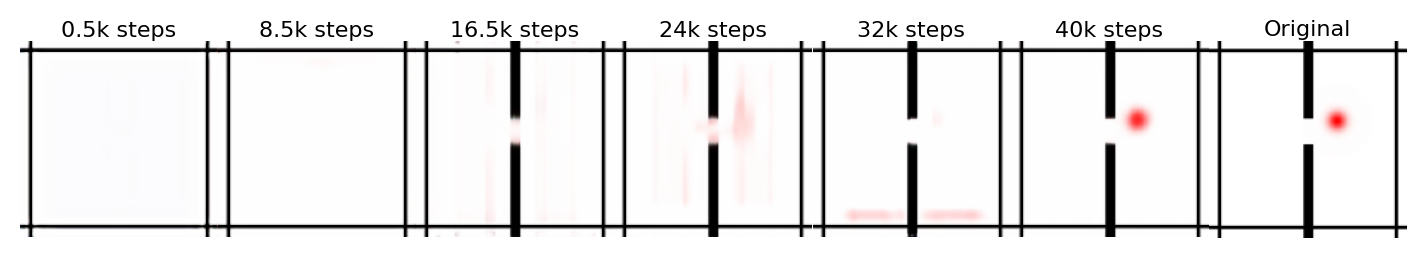} 
    \caption{Decoder training progress in TwoRoom. We track a single held-out example across training and show its reconstruction at multiple checkpoints; reconstructions become increasingly accurate over time. The ground-truth observation is shown on the right.}
    \label{fig:supp_tworoom_graded_examples3}
\end{figure}

\mypar{Kingdom}

\begin{figure}[H]
    \centering
    \includegraphics[width=\textwidth]{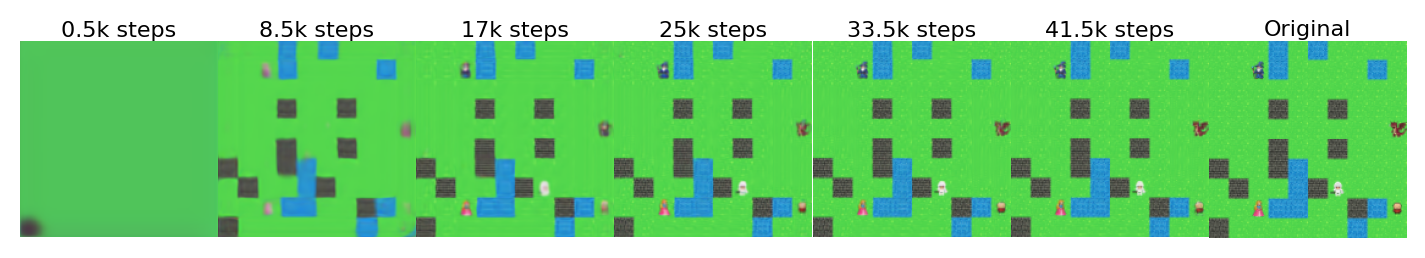} 
    \caption{Decoder training progress in Kingdom. We track a single held-out example across training and show its reconstruction at multiple checkpoints; reconstructions become increasingly accurate over time. The ground-truth observation is shown on the right.}
    \label{fig:supp_tworoom_graded_examples4}
\end{figure}

\mypar{OGBench Cube}

\begin{figure}[H]
    \centering
    \includegraphics[width=\textwidth]{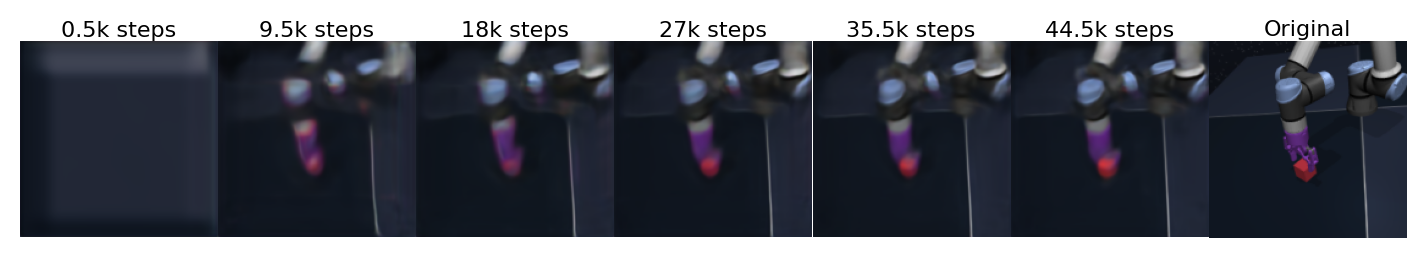} 
    \caption{Decoder training progress in OGBench Cube. We track a single held-out example across training and show its reconstruction at multiple checkpoints; reconstructions become increasingly accurate over time. The ground-truth observation is shown on the right.}
    \label{fig:supp_tworoom_graded_examples5}
\end{figure}

\subsection{Temporal Straightening}

\mypar{TwoRoom}

\begin{figure}[H]
    \centering
    \includegraphics[width=\textwidth]{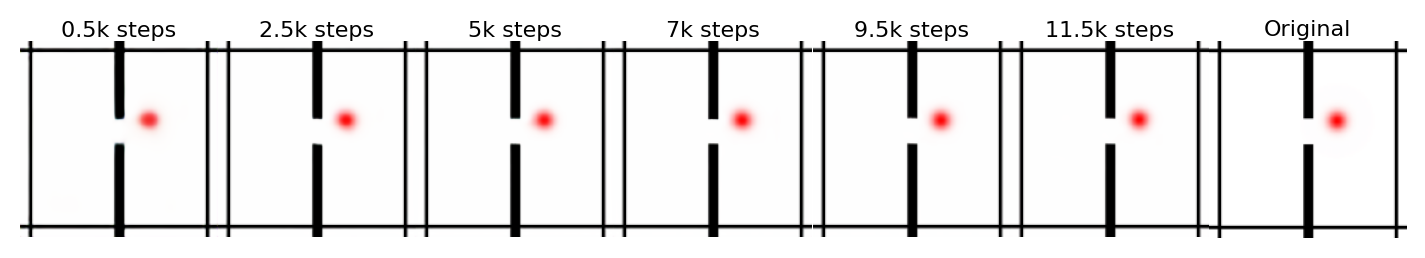} 
    \caption{Decoder training progress in TwoRoom. We track a single held-out example across training and show its reconstruction at multiple checkpoints; reconstructions become increasingly accurate over time. The ground-truth observation is shown on the right.}
    \label{fig:supp_tworoom_graded_examples6}
\end{figure}

\mypar{Kingdom}

\begin{figure}[H]
    \centering
    \includegraphics[width=\textwidth]{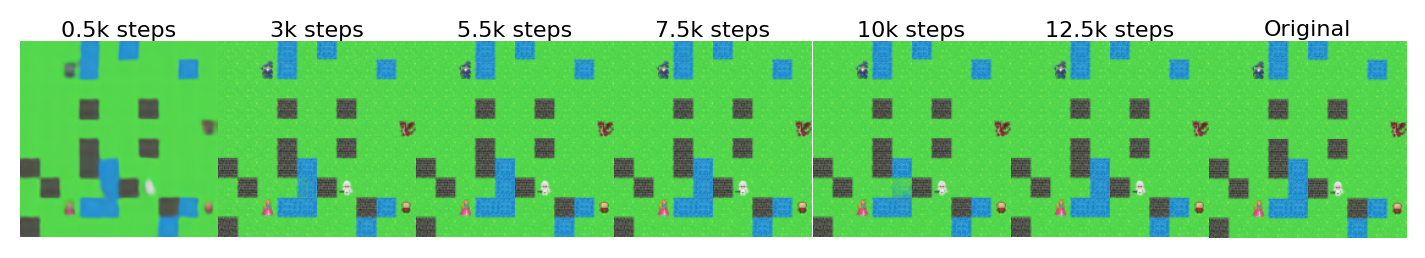} 
    \caption{Decoder training progress in Kingdom. We track a single held-out example across training and show its reconstruction at multiple checkpoints; reconstructions become increasingly accurate over time. The ground-truth observation is shown on the right.}
    \label{fig:supp_tworoom_graded_examples7}
\end{figure}

\mypar{OGBench Cube}

\begin{figure}[H]
    \centering
    \includegraphics[width=\textwidth]{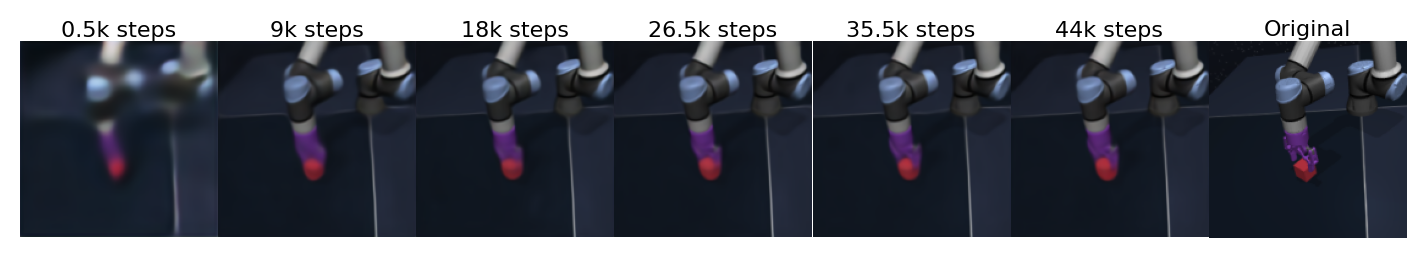} 
    \caption{Decoder training progress in OGBench Cube. We track a single held-out example across training and show its reconstruction at multiple checkpoints; reconstructions become increasingly accurate over time. The ground-truth observation is shown on the right.}
    \label{fig:supp_tworoom_graded_examples}
\end{figure}

\section{Robo-Dopamine evaluation}
\label{sec:robodopamine-voc}

While Robo-Dopamine's inference cost precludes its use in MPC-based planning, we assess the quality of its reward signal in isolation using the VOC (Value of Correlation) metric~\citep{ma2024visionlanguagemodelsincontext}. This metric measures the Spearman rank correlation between the predicted reward and task progress, offering a direct comparison of reward monotonicity independent of planning performance.

\begin{figure}[H]
    \centering
    \includegraphics[width=0.85\linewidth]{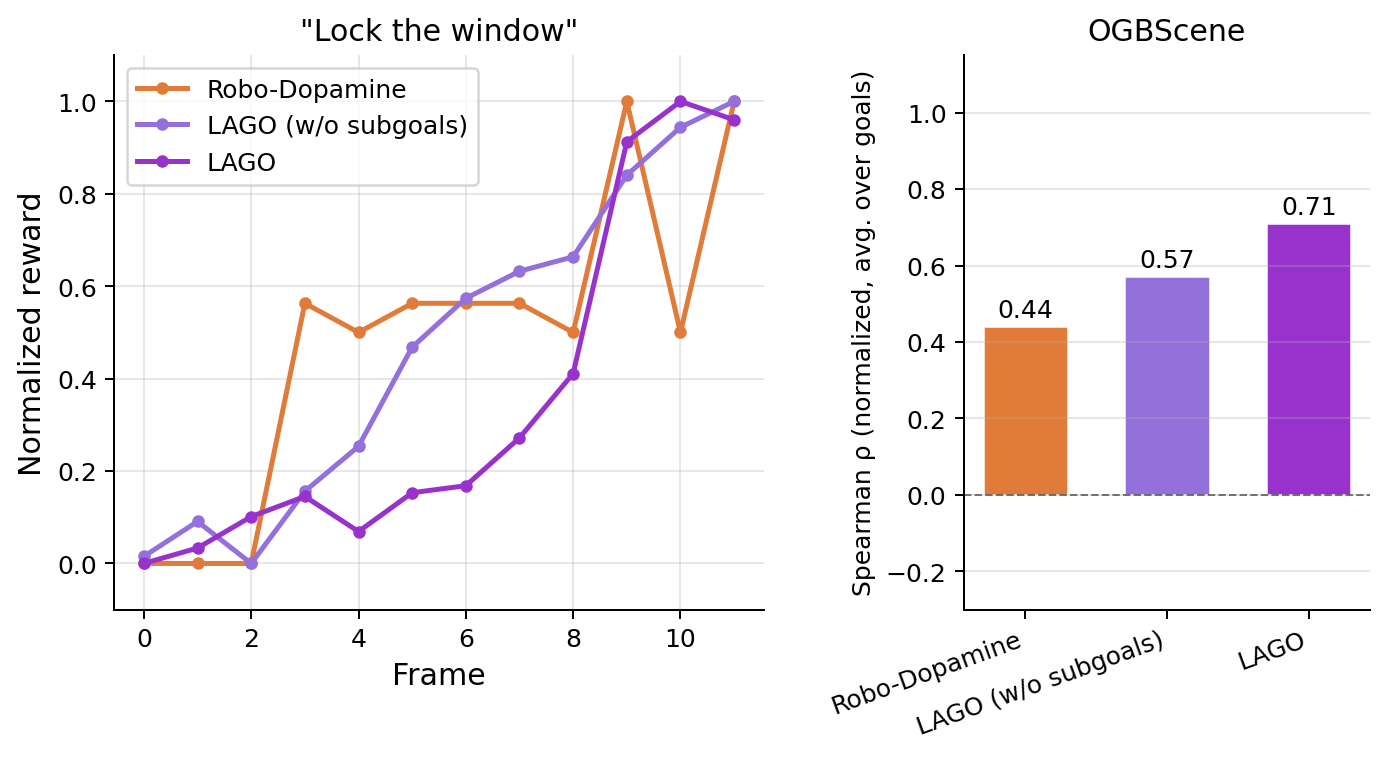}
    \caption{\textbf{Reward signal quality on OGBScene: LAGO vs.\ Robo-Dopamine.}
    \textit{Left:} Normalized reward over frames for the goal ``Lock the window''.
    Despite a broadly increasing trend for both, Robo-Dopamine ($\text{VOC}=0.69$) is noisier and assigns near-flat reward mid-task, while LAGO ($\text{VOC}=0.97$) rises more consistently.
    \textit{Right:} Average VOC across all OGBScene goals, where LAGO yields higher scores throughout.}
    \label{fig:robo-dopamine}
\end{figure}

As shown in \cref{fig:robo-dopamine}, LAGO produces a stronger and more monotonic reward signal than Robo-Dopamine across all OGBScene goals. On the representative goal ``Lock the window'', LAGO achieves a VOC of 0.97 versus 0.69 for Robo-Dopamine, whose reward remains near-flat through the middle of the episode and exhibits considerably more noise. This pattern holds on average across all goals, where LAGO consistently yields higher VOC scores. Taken together, these results suggest that LAGO not only satisfies the inference constraints of model-based planning, but also provides a higher-quality reward signal than a generative model with significantly more parameters.

\section{LAGOSET: Scaling Task Understanding Without Actions}
\label{app:LAGOSET}

LAGO learns dynamics from action-labeled transitions, but its subgoal predictor requires only video and language. To test whether the language-conditioned component benefits from data without actions, we train a LAGO model on interaction and demonstration data from LIBERO90 \citep{liu2023libero}, but augment the demonstration stream with LAGOSET, an action-less video-language corpus we assemble from a collection of larger demonstration datasets. We call the resulting model LAGO++, which differs from LAGO only in its subgoal supervision. We then evaluate whether scaling the language stream leads to increased robustness to out-of-domain instructions on held-out LIBERO-90 starting configurations.

\subsection{Dataset and Training}
\label{app:LAGOSET-data}

LAGOSET pools 310k language-annotated clips from BridgeData~V2~\citep{walke2023bridgedata}, DROID~\citep{khazatsky2025droidlargescaleinthewildrobot}, AgiBot World~\citep{agibotworldcontributors2025agibotworldcolosseolargescale}, and Language-Table~\citep{lynch2022interactivelanguagetalkingrobots}, where a clip is one camera view of one episode. The corpus spans about 165k episodes and 57.5M frames. Only video and instructions are used; actions are discarded. Language-Table clips retain their human annotations, while clips from the other sources are annotated with four instructions each, generated by an image-conditioned paraphraser (Qwen3.5-4B~\citep{qwen3.5}), of which one is sampled at each training visit. With LAGOSET, we simulate the difference in size between action-labeled robotic data and action-less demonstration data \citep{miech2019howto100mlearningtextvideoembedding,grauman2022ego4dworld3000hours,chen2026action100mlargescalevideoaction}.

\subsection{Rewording Benchmark}
\label{app:LAGOSET-eval}

The benchmark consists of 20 LLM-authored rewordings for each of 18 LIBERO-90 instructions, ranging from synonym substitutions to descriptions of the outcome only. Rewordings within cosine similarity $0.9$ (EmbeddingGemma) of any LAGOSET instruction are removed, leaving 289. Each rewording is assigned to one of four bins by its distance ($1-\cos$) to the original instruction; Table~\ref{tab:rewording-examples} gives examples.

Evaluation uses held-out initial configurations (five per task, 90 in total). Given a rewording, the model predicts the final subgoal $\tilde z_{\rho=1}$ in DINOv2 latent space, which is scored as correct if its nearest candidate in latent MSE is the final state of the instructed task. Candidates are the final and halfway states of all tasks in the same scene. Restricting candidates to the same scene keeps the test focused on instruction understanding, since final states of other scenes would be trivially far. The halfway states penalize predictions that stall part-way through the task. With about nine candidates per query (after filtering for leakage with LAGOSET as mentioned earlier), chance accuracy is $\approx 12\%$, while retrieving from the real final frame instead gives a ceiling of $100\%$. Accuracy is averaged over each instruction's rewordings and then over the 18 instructions.

\begin{table}[H]
  \caption{Example rewordings from each distance range, with their EmbeddingGemma distance to the original instruction.}
  \label{tab:rewording-examples}
  \small
  \centering
  \resizebox{\linewidth}{!}{%
  \begin{tabular}{lll}
    \toprule[\heavyrulewidth]
    Distance & ``put the black bowl on the plate'' & ``put the moka pot on the stove'' \\
    \midrule
    0.01--0.06 & place the black bowl on the plate (0.02) & place the moka pot on the stove (0.03) \\
    0.06--0.12 & put the black bowl on the dinner plate (0.06) & put the moka pot on the cooktop (0.07) \\
    0.12--0.19 & put the dark bowl on the dish (0.14) & put the espresso maker on the stove (0.16) \\
    0.19--0.50 & take the dark bowl and lower it onto the flat dish (0.22) & lift the espresso pot and rest it on the burner (0.30) \\
    \bottomrule
  \end{tabular}%
  }
\end{table}

\subsection{Results}
\label{app:LAGOSET-results}

\begin{table}[H]
  \caption{Final-subgoal retrieval accuracy (\%) on LIBERO-90 under instruction rewordings, binned by distance to the original instruction. Columns give trainable parameters; LAGO++ adds action-less LAGOSET language. Mean $\pm$ s.e.\ over instructions; higher is better.}
  \label{tab:rewording}
  \small
  \centering
  \resizebox{\linewidth}{!}{%
  \begin{tabular}{lccccc}
    \toprule[\heavyrulewidth]
    Distance & Rewordings & LAGO (15M) & LAGO++ (15M) & LAGO++ (30M) & LAGO++ (59M) \\
    \midrule
    Original & -- & 100.0\ppm{0.0} & 100.0\ppm{0.0} & 100.0\ppm{0.0} & 100.0\ppm{0.0} \\
    0.01--0.06 & 73 & 93.4\ppm{3.8} & 98.4\ppm{1.0} & 98.6\ppm{1.0} & 98.8\ppm{1.0} \\
    0.06--0.12 & 72 & 92.0\ppm{2.5} & 97.4\ppm{1.2} & 98.7\ppm{0.9} & 98.4\ppm{1.2} \\
    0.12--0.19 & 72 & 77.9\ppm{4.6} & 85.8\ppm{3.6} & 86.9\ppm{3.5} & 91.8\ppm{3.0} \\
    0.19--0.50 & 72 & 56.9\ppm{8.5} & 76.1\ppm{5.9} & 85.2\ppm{4.6} & 87.4\ppm{3.8} \\
    \midrule
    All & 289 & 79.2\ppm{4.4} & 89.7\ppm{2.3} & 91.5\ppm{1.9} & 94.6\ppm{1.3} \\
    \bottomrule
  \end{tabular}%
  }
\end{table}

Table~\ref{tab:rewording} reports accuracy per distance bin. Every model grounds the original instructions perfectly, so differences appear only under rewording. Accuracy falls as rewordings move away from the original, yet even LAGO, which has only seen the LIBERO-90 instructions, stays well above chance on the most distant rewordings ($56.9\%$ against $\approx 12\%$). LAGO++ (15M) has the same size as LAGO and differs only in its language data. LAGOSET contains no actions, and rewordings close to its instructions were filtered out, so its gains ($79.2\%$ to $89.7\%$ overall, largest on the most distant rewordings) show that LAGO can improve its task understanding from action-less demonstrations. Larger LAGO++ models improve further, again mostly on distant rewordings, so the most robust models combine action-less data with more capacity. Together, these results suggest that, on LIBERO-90, LAGO can use action-less language data to become more robust to rephrased instructions without additional action supervision.

\end{document}